\documentclass[conference]{IEEEtran}
\usepackage{times}

\usepackage[numbers]{natbib}
\usepackage{multicol}
\usepackage{hyperref}
\hypersetup{colorlinks=true,linkcolor=blue, linktocpage,urlcolor=blue}

\usepackage{amsmath}
\usepackage{amsfonts}
\usepackage{color}
\usepackage{graphicx}
\usepackage{hhline}
\usepackage{multirow}

\usepackage{subcaption}
\usepackage[font=footnotesize]{caption}

\usepackage[noend]{algpseudocode}
\usepackage{algorithm}

\newcommand\statespace{\mathbb{S}}
\newcommand\actionspace{\mathbb{A}}
\newcommand\goalspace{\mathbb{G}}
\newcommand\startstate{s_{0}}

\newcommand\approximateMDP{\hat{M}}

\newcommand\penalizedMDP{\tilde{M}}

\newcommand\incorrectset{\mathcal{X}}

\newcommand\buffer{\mathcal{D}}
\newcommand\reals{\mathbb{R}}

\newcommand\covering{\mathcal{C}}

\newtheorem{definition}{Definition}[section]
\newtheorem{assumption}{Assumption}[section]
\newtheorem{theorem}{Theorem}[section]

\makeatletter
\def\blfootnote{\xdef\@thefnmark{}\@footnotetext}
\makeatother

\begin{document}

\title{Planning and Execution using Inaccurate Models with Provable Guarantees}


\author{
  \authorblockN{Anirudh Vemula\authorrefmark{2}, Yash
    Oza\authorrefmark{2}, J. Andrew Bagnell\authorrefmark{3}
    and Maxim Likhachev\authorrefmark{2}}
  \authorblockA{\authorrefmark{2} Robotics Institute, Carnegie Mellon
    University \\
    \authorrefmark{3} Aurora Innovation \\
  \texttt{vemula@cmu.edu}}
}



%

\maketitle

\begin{abstract}
Models used in modern planning problems to simulate outcomes of real
world action executions are becoming increasingly complex,
ranging from simulators that do physics-based reasoning to precomputed
analytical motion primitives.
However, robots operating in the real
world often face situations not modeled by
these models before execution.
This imperfect modeling can lead to highly
suboptimal or even incomplete behavior during execution.
In this paper, we propose \textsc{Cmax} an approach for interleaving planning and
execution. \textsc{Cmax} adapts its planning strategy online during
real-world execution to account for any
discrepancies in dynamics during planning,
without requiring updates to the dynamics of the model.
This is achieved by biasing the planner away from transitions whose
dynamics are discovered to be inaccurately modeled, thereby leading to robot behavior
that tries to complete the task despite having
an inaccurate model.
We provide provable guarantees on the
completeness and efficiency of the proposed planning and execution framework under specific
assumptions on the model, for both small and large state
spaces.
Our approach \textsc{Cmax} is shown to be efficient empirically in simulated robotic
tasks including 4D planar pushing, and in real
robotic experiments using PR2 involving a 3D pick-and-place task where the mass
of the object is incorrectly modeled, and a 7D
arm planning task where one of the joints is not operational
leading to discrepancy in dynamics. The video of our physical robot experiments can be found at
\url{https://youtu.be/eQmAeWIhjO8}.\blfootnote{A blog post summarizing
  this work can found at \url{https://vvanirudh.github.io/blog/cmax/}}
\end{abstract}

\IEEEpeerreviewmaketitle

\section{Introduction}
\label{sec:introduction}

Modern robotic planning approaches involve use of models that tend to
be sophisticated and complex. These models are used to simulate the
dynamics of the real world and foresee the outcomes of actions
executed. From using fast analytical solvers to generate motion
primitives on-the-fly \cite{DBLP:conf/icra/CohenSCL11} to 
simulators that do reasoning based on physics, and optimization to resolve
contacts \cite{DBLP:conf/iros/TodorovET12}, these models are getting better at
modeling the dynamics of the real world. However, real world robotic
tasks are rife with situations that cannot be predicted and therefore,
modeled before execution. Thus, we need a
planning approach that can use potentially inaccurate models and still
complete the task.

\begin{figure}[t]
  \centering
  \begin{subfigure}{.55\columnwidth}
    \includegraphics[width=\linewidth]{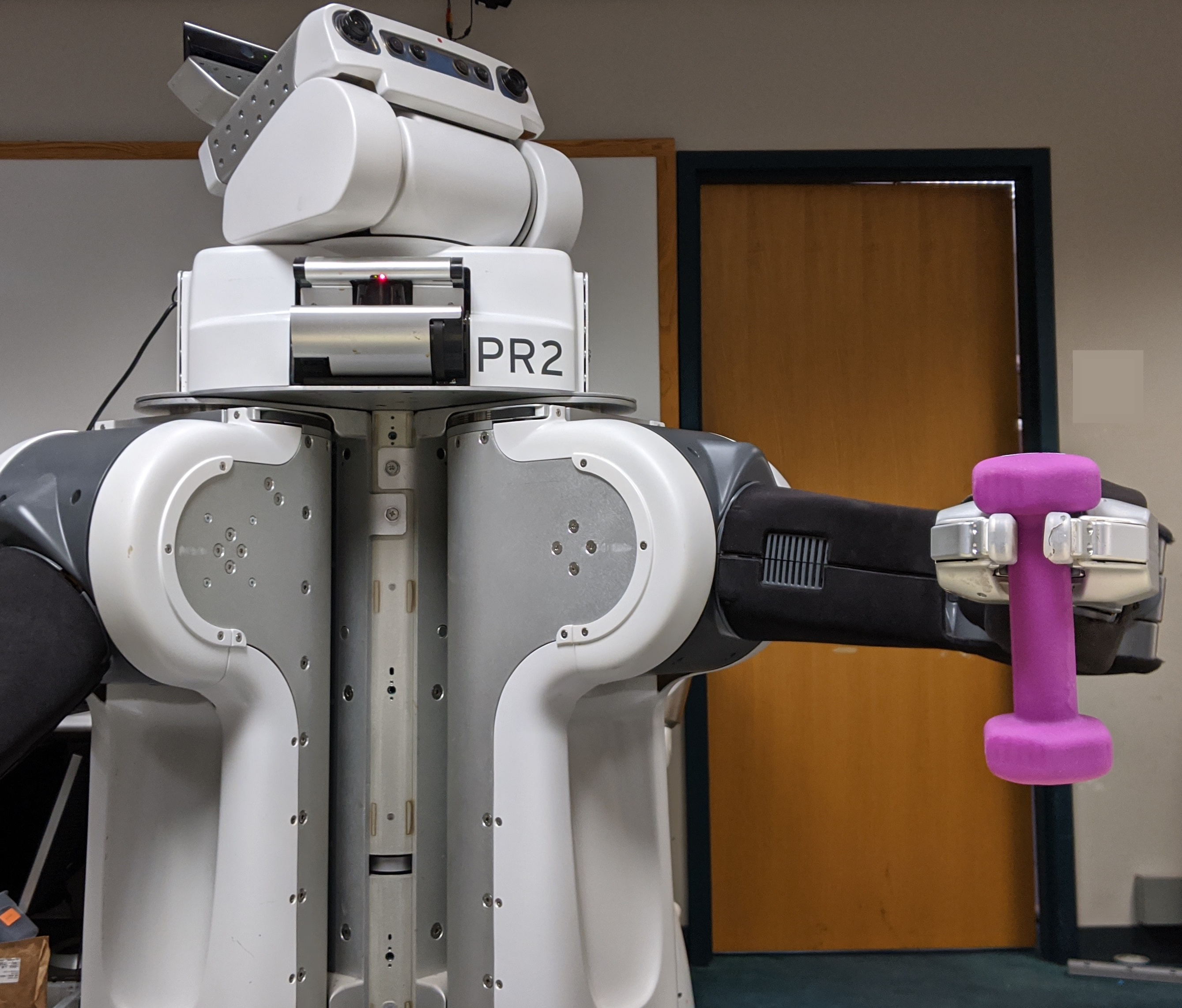}
  \end{subfigure}
  \begin{subfigure}{.42\columnwidth}
    \includegraphics[width=\linewidth]{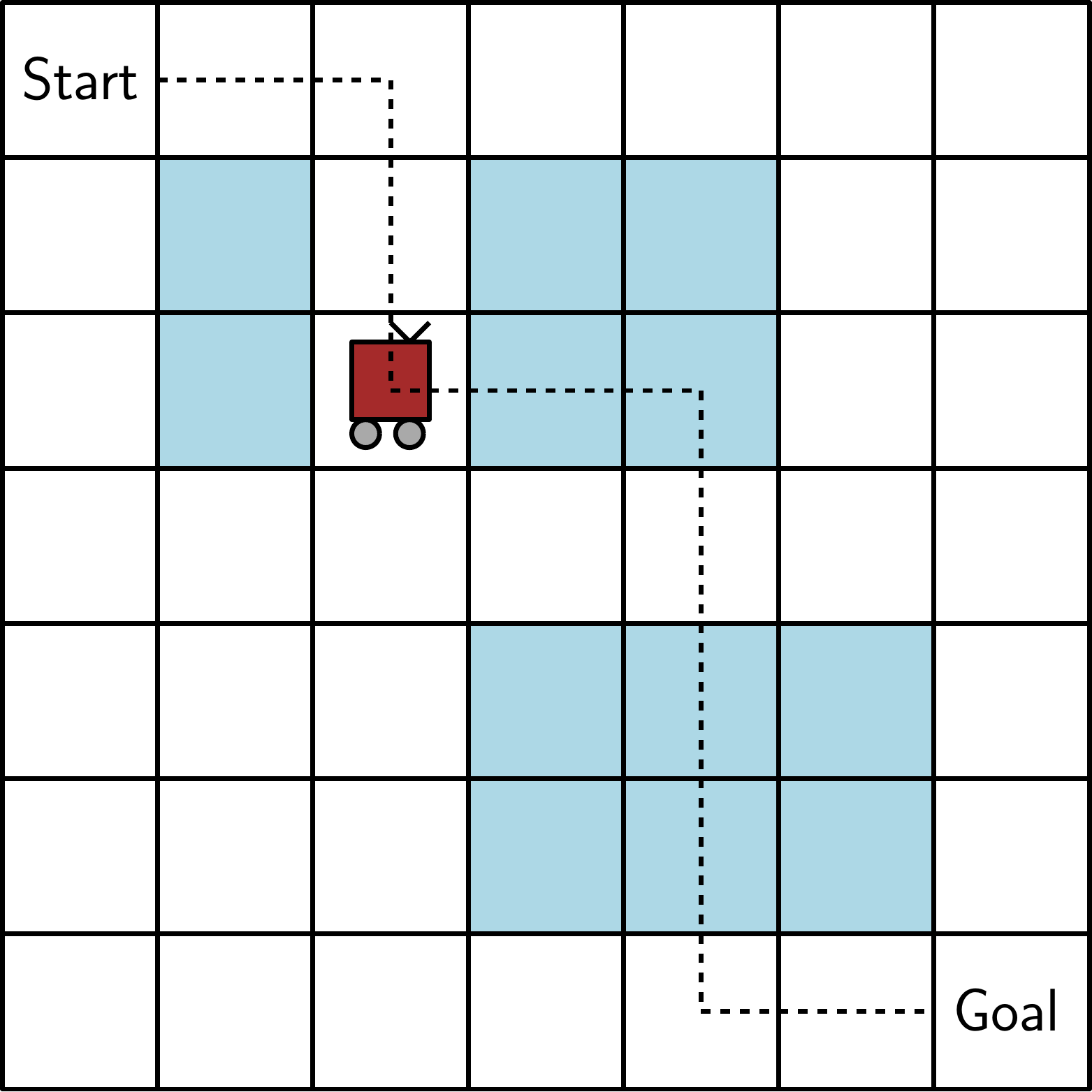}
  \end{subfigure}  
  \caption{(left) PR2 executing a pick-and-place task with a heavy
    object that is modeled as light, resulting in hitting joint torque
    limits during execution. (right) Mobile
    robot navigating a gridworld with icy states, where the robot slips, that are not modeled
    as icy resulting in discrepancy in dynamics.}
  \label{fig:intro}
  \vspace{-0.5cm}
\end{figure}

For example, consider the task depicted in Figure~\ref{fig:intro}
(left) where a robotic arm needs to pick an object and place it at a
goal location. Without knowledge of the mass of the object, the model
can be inaccurate in simulating the dynamics. If the object is modeled
as light, the planned 
path would pick it to a certain height before placing it at the
goal location. However, if the object is heavy in the real world, like
in Figure~\ref{fig:intro} (left), this plan cannot be executed as the
joint torque limits are reached and the arm cannot move higher. Thus, by
using the inaccurate model for planning, the arm is stuck and cannot reach the
goal. Figure~\ref{fig:intro} (right) presents another simple scenario
where a mobile robot is navigating a gridworld containing icy
states, where the robot slips, i.e.
if the robot tries to go right or left in an icy state, it will move two cells rather
than one cell in that direction. However, the model used for planning
does not model the icy states and hence, cannot
simulate the real world dynamics correctly. This can lead to highly
suboptimal paths or sometimes even inability to reach the goal, when using such a model for planning.

A typical solution to this problem is to update the dynamics of the
model and replan \cite{DBLP:journals/sigart/Sutton91}. However, this
is often impossible in real world planning problems 
where we use models that are complex and in some cases obtained from
expensive computation that is done offline before execution
\cite{DBLP:conf/wafr/HauserBHL06}. The
dynamics of these models cannot be changed online arbitrarily without
deteriorating their simulation capabilities in other scenarios and sacrificing
real-time execution. In addition, this solution might require us to have
the knowledge of what part of the model dynamics is inaccurate and
how to correct it. Going
back to the pick-and-place example in Figure~\ref{fig:intro}, to
update the model we need to first identify
that the modeled mass is incorrect and then estimate the true mass to
correct the dynamics of the model. Both of these steps require
specialized non-trivial implementations. Finally, in the case of models that
\emph{can} be updated online efficiently, it might still not be possible to
model the true dynamics without an unreasonably large number of online
executions because the true dynamics are
often very complex, e.g. modeling cooperative navigation dynamics in
human crowds \cite{DBLP:conf/icra/VemulaMO17}. The above aspects make the 
solution of updating model dynamics online undesirable in real world robotic
tasks, where we are interested in completing the task and \emph{not} in
modeling the dynamics accurately.

In this work, we present an alternative approach \textsc{Cmax} for
interleaving planning and 
execution that does not require updating
the dynamics of the model. Instead during execution, whenever we discover an action
where the dynamics differ between the real world and the model, we
update the cost function to penalize executing such state-action pairs
in the future. This biases the planner to replan paths that do not
consist of such state-action pairs, and thereby avoid regions of
state-action space where the dynamics are known to differ. Based on
this idea, we present
algorithms for both small state spaces, where we can do
exact planning, and large state spaces, including continuous state
spaces, where we resort to function 
approximation to update the cost function and to maintain cost-to-go
estimates.
Our framework \textsc{Cmax} comes with provable guarantees on
reaching 
the goal, without any resets, under specific assumptions on
the model.
The proposed algorithms are tested on a range of tasks
including simulated 4D planar pushing as well as
physical robot 3D pick-and-place task where the mass of the object is
incorrectly modeled, and 7D arm planning tasks when one of the joints
is not operational, leading to discrepancy in dynamics.

\section{Preliminaries}
\label{sec:preliminaries}

We are interested in the deterministic shortest path problem
represented by the tuple $M = (\statespace, \actionspace, \goalspace,
f, c)$ where $\statespace$ denotes the state space, $\actionspace$
denotes the action space, $\goalspace \subseteq \statespace$ is the
non-empty set of goal states we are
interested in reaching, $f: \statespace \times \actionspace
\rightarrow \statespace$ denotes the deterministic dynamics governing
the transition to next state given current state and action, and $c:
\statespace \times \actionspace \rightarrow [0, 1]$ is the cost
function. For the purposes of
this work, we will focus on small
discrete action spaces, bounded costs lying between $0$ and
$1$\footnote{Any bounded non-negative cost can be scaled to fit
  this assumption}, and a cost-free termination goal state i.e. for
all $g \in \goalspace$, we have
$c(g, a) = 0$
and $f(g, a) = g$ for all actions $a \in \actionspace$.
The objective of the shortest path problem is to find the least-cost path from any given start state $\startstate
\in \statespace$ to any goal state $g \in \goalspace$
in $M$.
We assume that there
exists at least one path from each state $s \in
\statespace$ to one of the goal states $g \in \goalspace$ in $M$, and
that the
cost of any transition starting from a non-goal state is positive i.e. $c(s, a)
> 0$ for all $s \in \statespace \setminus \goalspace, a \in
\actionspace$. These assumptions are typical for analysis in
deterministic shortest path problems
\cite{DBLP:books/lib/Bertsekas05}. 
We use $V(s)$ to denote the cost-to-go estimate of any state $s \in
\statespace$ and $V^*(s)$ to denote the optimal cost-to-go. From
dynamic programming literature \cite{DBLP:books/lib/Bertsekas05}, we know
that the optimal cost-to-go satisfies the Bellman optimality condition
$V^*(s) = \min_{a \in \actionspace} [c(s, a) + V^*(f(s, a))]$.
A cost-to-go estimate $V$ is
called admissible if it underestimates the optimal cost-to-go
$V(s) \leq V^*(s)$ for all $s \in \statespace$, and
is called consistent if it satisfies the condition that for any
state-action pair $(s, a), s\notin\goalspace$, $V(s) \leq c(s, a) + V(f(s, a))$, and $V(g) = 0$ for
all $g \in \goalspace$.

In this work, we assume that the exact dynamics are
initially unknown to the robot, and can only be discovered through
executions. Thus, instead of offline planning
methods, we need
online methods that interleave planning with action execution. Specifically, we 
focus on the online real-time planning
setting where the robot does not have access to resets, and the
robot has to interleave planning and execution to ensure real-time
operation. This is similar to the classical real-time search setting
considered by works like LRTA*~\cite{DBLP:journals/ai/Korf90},
RTAA*~\cite{DBLP:conf/atal/KoenigL06}, RTDP~\cite{DBLP:journals/ai/BartoBS95} and
several others. An important aspect of these approaches is that the robot can only perform a fixed amount of
computation for planning, independent of the size of state space, before it has to execute an action.

\section{Problem Setup}
\label{sec:problem-setup}

Consider the problem of a robot acting to find a least-cost path to a
goal in an environment represented
by the tuple $M = (\statespace, \actionspace, \goalspace, f, c)$ with unknown
deterministic dynamics $f$ and known cost function $c$. The robot
gathers knowledge of the dynamics over a single
trajectory in the environment, and does not have access to any
resets, ruling out any episodic approach.
This is an extremely challenging setting as the robot has to reason
about whether to exploit its current knowledge of the dynamics to act
near-optimally or to explore to gain more knowledge of the dynamics,
possibly at the expense of suboptimality.

We assume that the agent has access to an approximate model, $\hat{M} = (\statespace,
\actionspace, \goalspace, \hat{f}, c)$, that it can use to simulate the outcome of
its actions and use for planning. In our motivating gridworld example
(Figure~\ref{fig:intro} right), this model
represents a grid with no icy states, so the dynamics $\hat{f}$
moves the robot to the next cell based on the executed
action without any slip. However, the real environment contains
icy states
resulting in dynamics $f$ that differ on state-action pairs where the
state is icy. For the remainder of this paper, we will
refer to such state-action pairs where $f$ and $\hat{f}$ differ as ``incorrect''
state-action pairs, and 
use the notation $\incorrectset \subseteq
\statespace \times \actionspace$ to denote the set of ``incorrect'' state-action
pairs,
i.e. $f(s, a) \neq \hat{f}(s, a)$ for all $(s, a) \in
\incorrectset$.
The objective is for the robot to reach a goal
state from a given start state, despite using an inaccurate model for planning, while
minimizing the cost incurred and ensuring real-time execution.


\section{Approach}
\label{sec:approach-1}

Existing planning and learning approaches try to learn a very good
approximation of $M$ from scratch 
through online executions \cite{DBLP:journals/ml/KearnsS02, DBLP:journals/jmlr/BrafmanT02, DBLP:conf/atal/JongS07,
  DBLP:journals/pami/DeisenrothFR15}, or update the dynamics of model
$\hat{M}$ so that it approximates $M$ well \cite{DBLP:conf/icml/AbbeelQN06,
  DBLP:conf/aaai/Jiang18, rastogi2018sample}.
In this work, we propose an approach \textsc{Cmax} that uses
the inaccurate model $\hat{M}$ online \emph{without} updating its dynamics, and
is provably guaranteed to complete the task.
In a nutshell, instead of learning a new dynamics model from scratch or
updating the dynamics of existing model,
\textsc{Cmax}
maintains a
running estimate of the set $\incorrectset_t$ consisting of all
state-action pairs that have been executed and have been discovered to
be incorrect until timestep $t$. Using the set $\incorrectset_t$, we update the cost
function to bias the planner to plan future paths that avoid
state-action pairs that are known to be incorrect.
It is
important to note that the challenge of dealing with
exploration-exploitation dilemma online still exists, as we do not
know the set of state-action pairs $\incorrectset$ where the dynamics differ ahead of online
execution. A similar approach was proposed in \citet{DBLP:conf/aaai/Jiang18} for the
episodic setting where the robot had access to resets, and for small
state spaces where we could perform full state space planning.
\textsc{Cmax}
extends it to the significantly more challenging 
online real-time setting and we present a practical algorithm for
large state spaces. 

\subsection{Penalized Model}
\label{sec:penalized-model}

We formalize
our approach
as follows: Given a model
$\approximateMDP$ and a set $\incorrectset \subseteq \statespace
\times \actionspace$ consisting of state-action pairs that have been
discovered to be incorrect so far, define the penalized model $\penalizedMDP_\incorrectset$ as:
\begin{definition}[Penalized Model]
  The penalized model $\penalizedMDP_\incorrectset = (\statespace,
  \actionspace, \goalspace,
  \hat{f}, \tilde{c}_\incorrectset)$ has the same state space,
  action space, set of goals, and dynamics as $\approximateMDP$. The
  cost function $\tilde{c}_\incorrectset$ though
  is defined as $\tilde{c}_\incorrectset(s, a) = |\statespace|$ if $(s, a) \in
  \incorrectset$, else $\tilde{c}_\incorrectset(s, a) =
  c(s,a)$.\footnote{This is similar to the notion of penalized MDP,
    introduced in \citet{DBLP:conf/aaai/Jiang18}}
  \label{def:penalized-mdp}
\end{definition}

Intuitively, the penalized model $\penalizedMDP_{\incorrectset}$
has a very high cost for any transition where the dynamics differ,
i.e. $(s, a) \in \incorrectset$, and
the same cost as the model $\hat{M}$ otherwise. More specifically, the
cost is inflated to the size of the statespace, which is the maximum
cost of a path that visits all states\footnote{Hence, the name \textsc{Cmax} for our approach} (remember, that our cost is
normalized to lie within $0$ and $1$.) This biases the planner to ``explore'' all other
state-action pairs that are not yet known to be incorrect before it
plans a path through an incorrect state-action pair.
In the next section, we
will describe how we use the penalized model
$\penalizedMDP_\incorrectset$ for real-time planning.


\subsection{Limited-Expansion Search for Planning}
\label{sec:limit-expans-search}

During online execution, the robot has to constantly plan the next
action to execute from its current state in real-time. This forces the
robot to use a fixed amount of computation for planning before it has
to execute the best action found so far. In this work, we use
a real-time search method that is adapted from RTAA*
proposed by \citet{DBLP:conf/atal/KoenigL06}.

\begin{algorithm}[t]
  \caption{Limited-Expansion Search based on
    RTAA*\cite{DBLP:conf/atal/KoenigL06}}
  {\normalsize
  \begin{algorithmic}[1]
    \Function{$\mathtt{SEARCH}$}{$s,
      \penalizedMDP_{\incorrectset}, V, K$}
    \State Initialize $g(s) \leftarrow 0$
    \State Initialize min-priority open list $O$, and closed list $C$
    \State Add $s$ to open list $O$ with priority $g(s) + V(s)$
    \For{$i=1, 2, \cdots, K$}
    \State Pop $s_i$ from open list $O$
    \State If $s_i \in \goalspace$, then $s_{\mathsf{best}} \leftarrow
    s_i$ and move to Line~\ref{line:updates}
    \For{$a \in \actionspace$} \Comment{\textit{Expanding state $s_i$}}
    \State Get successor $s' = \hat{f}(s_i, a)$
    \State If $s' \in C$, continue to next action
    \If{$s' \in O$ and $g(s') > g(s_i) + \tilde{c}_\incorrectset(s_i, a)$}
    \State Update $g(s') \leftarrow g(s_i) + \tilde{c}_\incorrectset(s_i,
    a)$
    \State Reorder open list $O$
    \ElsIf{$s' \notin O$}
    \State Set $g(s') \leftarrow g(s_i) + \tilde{c}_\incorrectset(s_i, a)$
    \State Add $s'$ to $O$ with priority $g(s') + V(s')$
    \EndIf
    \EndFor
    \State Add $s_i$ to the closed list $C$
    \EndFor
    \State Pop $s_{\mathsf{best}}$ from open list
    $O$\label{line:pop-best-node}
    \For{$s' \in C$}\label{line:updates}
    \State Update $V(s') \leftarrow
    g(s_{\mathsf{best}}) + V(s_{\mathsf{best}}) - g(s')$\label{line:cost-to-go-update}
    \EndFor
    \State Backtrack from $s_{\mathsf{best}}$ to $s$, and set
    $a_{\mathsf{best}}$ as the first action on path from $s$ to
    $s_{\mathsf{best}}$
    
    \Return{$a_{\mathsf{best}}$}
    \EndFunction
  \end{algorithmic}}
  \label{alg:limited-expansion-search}
\end{algorithm}

The planner is summarized in Algorithm~\ref{alg:limited-expansion-search}. At any
timestep $t$, given the
current penalized model $\penalizedMDP_{\incorrectset_t}$ and the current
state $s_t$, the planner constructs a lookahead
search tree using at most $K$ state expansions. We obtain the successors of any
expanded state and the cost of any state-action pair using the
penalized model $\penalizedMDP_{\incorrectset_t}$.
After expanding $K$
states, it finds the best state $s_{\mathsf{best}}$ among the leaves of the search tree
that has the least sum of cost-to-come from $s_t$ and
cost-to-go to a goal state (line~\ref{line:pop-best-node} in Algorithm~\ref{alg:limited-expansion-search}). The best action to execute in the current
state $s_t$ is chosen to be the first action on the path from $s_t$ to
$s_{\mathsf{best}}$ in the search tree and the cost-to-go estimates of
all expanded states are updated as: $V(s_{\mathsf{expanded}}) =
g(s_{\mathsf{best}}) + V(s_{\mathsf{best}}) -
g(s_{\mathsf{expanded}})$, where $g(s)$ is the cost-to-come from $s_t$
for any state $s$ in the search tree.
The amount of computation used to compute the best action for the
current state is bounded as a factor of the number of expansions $K$ in the search
tree. Thus, we can bound the planning time and ensure real-time
operation for our robot.



\subsection{Warm Up: Small State Spaces}
\label{sec:small-state-spaces}

In this section, we will present an algorithm that is applicable for
small discrete state spaces where it is feasible to maintain cost-to-go
estimates for all states $s \in \statespace$ using a tabular representation, and we can maintain a
running set $\incorrectset_t$ containing all the discovered incorrect state-action pairs
so far, without resorting to function approximation. The algorithm\footnote{A similar algorithm in the
  episodic setting with full state space planning is presented in \citet{DBLP:conf/aaai/Jiang18}} is shown in Algorithm
\ref{alg:small-state-spaces}.
Intuitively, Algorithm \ref{alg:small-state-spaces} maintains a
running set of incorrect state-action pairs
$\incorrectset_t$, updates the set whenever it encounters an incorrect
state-action pair, and recomputes the penalized model
$\penalizedMDP_{\incorrectset_t}$. Crucially, the algorithm never updates the
dynamics of the model $\approximateMDP$, and only updates the cost
function according to Definition~\ref{def:penalized-mdp}.
In order to prove completeness, we assume the following:
\begin{assumption}
  Given a penalized model $\tilde{M}_{\incorrectset_t}$ and the current
  state $s_t$ at any timestep $t$, there always exists at least one path from $s_t$ to a goal
  state that \textit{does not contain} any state-action pairs $(s, a)$ that are known to
  be incorrect, i.e. $(s, a) \in \incorrectset_t$. \footnote{This
    assumption is less restrictive than the assumption that
    there exists at least one path from the current state to a goal
    that does not contain any state-action pairs $(s, a)$ that are incorrect i.e. $(s,
    a) \in \incorrectset$}
  \label{assumption:core}
\end{assumption}

\begin{algorithm}[t]
  \caption{\textsc{Cmax} -- Small State Spaces}
  {\normalsize
  \begin{algorithmic}[1]
    \State Initialize $\approximateMDP_1 \leftarrow \approximateMDP$,
    $\incorrectset_1 \leftarrow \{\}$, start state $s_1 \in
    \statespace$, cost-to-go estimates $V$, number of expansions $K$,
    $t \leftarrow 1$
    \While{$s_t \notin \goalspace$}
    \State Get $a_t = \mathtt{SEARCH}(s_t, \hat{M}_t, V, K)$
    \State Execute $a_t$ in environment $M$ to get $s_{t+1} = f(s_t, a_t)$
    \If{$s_{t+1} \neq \hat{f}(s_t, a_t)$}
    \State Add $(s_t, a_t)$ to the set : $\incorrectset_{t+1} \leftarrow \incorrectset_t \cup
    \{(s_t, a_t)\}$
    \State Update the penalized model : $\approximateMDP_{t+1} \leftarrow
    \penalizedMDP_{\incorrectset_{t+1}}$
    \Else
    \State $\incorrectset_{t+1} \leftarrow \incorrectset_t$,
    $\approximateMDP_{t+1} \leftarrow \approximateMDP_t$
    \EndIf
    \State $t \leftarrow t + 1$
    \EndWhile
  \end{algorithmic}}
  \label{alg:small-state-spaces}
\end{algorithm}

Under this assumption, we can show
the following guarantee for
Algorithm \ref{alg:small-state-spaces}:
\begin{theorem}
  Assume Assumption~\ref{assumption:core} holds then, if $\incorrectset$ denotes
  the set consisting of all incorrect state-action
  pairs, and the initial cost-to-go estimates used are
  admissible and consistent, then using Algorithm~\ref{alg:small-state-spaces}
  the robot is guaranteed to reach a goal state in at most
  $|\statespace|^2$ timesteps. Furthermore, if we allow for $K =
  |\statespace|$ expansions, then we can guarantee that the
  robot will reach a goal state
  in at most $|\statespace|(|\incorrectset|+1)$ timesteps.
  \label{thm:small-state-spaces}
\end{theorem}
Proof of the above theorem is given in Appendix
\ref{sec:proof-theor-refthm:s}.
The above
theorem establishes that using Algorithm 
\ref{alg:small-state-spaces}, the robot is guaranteed to reach a goal
state under Assumption~\ref{assumption:core}. In
practice, we observe that the number of timesteps to reach a goal has a smaller
dependence on the size of state space than the
worst-case bound, especially if Algorithm~\ref{alg:small-state-spaces}
starts with cost-to-go estimates that are reasonably accurate for the
initial model $\hat{M}$. 

\subsection{Large State Spaces}
\label{sec:large-state-spaces}

In large state spaces, it is infeasible to
maintain cost-to-go estimates for all states $s \in \statespace$ using
a tabular representation and maintain a running estimate of the set
$\incorrectset_t$, as both
could be very large in size. Thus,
we will need to resort to function approximations for both
cost-to-go estimates and the set $\incorrectset_t$.

We will assume existence of a fixed distance metric $d:
\statespace \times \statespace \rightarrow \mathbb{R}^+\cup\{0\}$, and that
$\statespace$ is bounded under this metric.
We relax the definition of $\incorrectset$ using the distance
metric $d$ as follows: Define any state-action pair $(s, a) \in
\incorrectset^\xi$ to be $\xi$-incorrect if $d(f(s, a), \hat{f}(s, a))
> \xi$ where $\xi \geq  0$. We assume that there is an underlying
path following controller that is used to execute our plan and can deal with
discrepancies smaller than $\xi$. Thus, we allow for small
discrepancies in our approximate model $\hat{M}$ that can be resolved
using a low-level controller.

Our algorithm for  large state
spaces is presented in Algorithm \ref{alg:large-state-spaces}. The main idea
of the algorithm is to ``cover'' the set $\incorrectset^\xi$ using
hyperspheres in $\statespace\times\actionspace$. Since the action
space $\actionspace$ is a discrete set, we maintain separate sets of
hyperspheres for each action $a \in \actionspace$. Whenever the agent encounters an incorrect
state-action pair $(s, a) \in \incorrectset^\xi$, it places a
hypersphere at $s$ corresponding to action $a$ whose radius (as
measured by the metric $d$) is given
by $\delta > 0$, a domain-dependent constant.
We inflate the cost of a
state-action pair $(s, a)$, according to
Definition~\ref{def:penalized-mdp},
if $s$ lies inside any hypersphere corresponding to action $a$. In
practice, this
is implemented by constructing 
separate KD-Trees in state space $\statespace$ for each action
$a \in \actionspace$ to enable efficient lookup.

After executing the action and placing a hypersphere if a discrepancy
in dynamics
was observed, the function approximation for cost-to-go is updated
iteratively as follows (Line~\ref{line:iteration-start} to
Line~\ref{line:iteration-end}): Sample a batch of states from the buffer
of previously visited states with replacement, construct a lookahead
tree for each state in the batch (through parallel jobs) to obtain all
states on the closed list and their corresponding cost-to-go updates
using Algorithm~\ref{alg:limited-expansion-search},
and finally update the parameters of the cost-to-go function
approximator to minimize the mean squared loss $\mathcal{L}(V_\theta, \mathbb{X}) = \frac{1}{2|\mathbb{X}|} \sum_{(s,
    V(s)) \in \mathbb{X}} (V(s) - V_\theta(s))^2$ for all the expanded
states through a gradient descent step (Line~\ref{line:iteration-end}).

Observe that, similar to Algorithm~\ref{alg:small-state-spaces}, we
do not update the dynamics $\hat{f}$ of the model, and only update the
cost function according to
Definition~\ref{def:penalized-mdp}. However, unlike
Algorithm~\ref{alg:small-state-spaces}, we do not explicitly maintain
a set of incorrect state-action pairs but maintain it implictly
through hyperspheres. By using hyperspheres, we obtain local
generalization and increase the cost of all the state-action pairs
inside a hypersphere.
In addition, unlike
Algorithm~\ref{alg:small-state-spaces}, we update cost-to-go estimates
of not only the expanded states in the lookahead tree obtained from
current state $s_t$, but also from previously visited states. This
ensures that the function approximation used for maintaining
cost-to-go estimates does not deteriorate for states that were
previously visited, and potentially help in generalization.
\begin{algorithm}
  \caption{\textsc{Cmax} -- Large State Spaces}
  {\normalsize
  \begin{algorithmic}[1]
    \State Initialize $\approximateMDP_1 \leftarrow \approximateMDP$,
    Cost-to-go function approximation $V_{\theta_1}$, Set of
    hyperspheres $\incorrectset^\xi_1 \leftarrow \{\}$, Start state
    $s_1$, Number of planning updates $N$, Batch size $B$, Buffer
    $\buffer$, Number of expansions $K$,
    Learning rate $\eta$, $t \leftarrow 1$, Radius of hypersphere
    $\delta$, Discrepancy threshold $\xi$
    \While{$s_t \notin \goalspace$}
    \State Get $a_t \leftarrow \mathtt{SEARCH}(s_t, \approximateMDP_t,
    V_{\theta_{t}}, K)$
    \State Execute $a_t$ in environment $M$ to get $s_{t+1} \leftarrow
    f(s_t, a_t)$
    \If{$d(s_{t+1}, \hat{f}(s_t, a_t)) > \xi$}
    \State Add $\incorrectset_{t+1}^\xi \leftarrow
    \incorrectset_t^\xi \cup \{\mathsf{sphere}(s_t, a_t, \delta)\}$
    \Else
    \State $\incorrectset_{t+1}^\xi \leftarrow \incorrectset_t^\xi$
    \EndIf
    \State Update $\approximateMDP_{t+1} \leftarrow
    \tilde{M}_{\incorrectset_{t+1}^\xi}$
    \State Add $s_t$ to buffer $\buffer$
    \State Update $V_{\theta_{t+1}} \leftarrow
    \mathtt{UPDATE}(s_t, \approximateMDP_{t+1}, V_{\theta_t}, \buffer)$
    \State $t \leftarrow t + 1$
    \EndWhile

    \Function{$\mathtt{UPDATE}$}{$s, \approximateMDP, V_\theta,
      \buffer$}
    \For{$n=1, \cdots, N$}
    \State Sample batch of $B$ states $S_n$ from buffer $\buffer$
    with replacement\label{line:iteration-start}
    \State Call $\mathtt{SEARCH}(s_i, \hat{M}, V_\theta, K)$ for each
    $s_i \in S_n$ to get all states on closed list $s_i'$ and their corresponding cost-to-go updates $V(s_i')$
    and construct the training set $\mathbb{X}_n = \{(s_i', V(s_i'))\}$
    \State Update: $\theta \leftarrow \theta - \eta\nabla_\theta
    \mathcal{L}(V_\theta, \mathbb{X}_n)$\label{line:iteration-end}
    \EndFor
    \Return{$V_\theta$}
    \EndFunction
  \end{algorithmic}}
\label{alg:large-state-spaces}
\end{algorithm}

We can provide a guarantee on the completeness of
Algorithm~\ref{alg:large-state-spaces} by assuming the following: 
\begin{assumption}
  Given a penalized model $\penalizedMDP_{\incorrectset_t^\xi}$ and
  the current state $s_t$ at any timestep $t$ during execution, there
  always exists at least one path from $s_t$ to a goal state that is
  \textit{at least $\delta$ distance away} from any state-action pair $(s, a)$ 
  that is known to be $\xi$-incorrect, i.e. $(s, a) \in \incorrectset_t^\xi$.
  \label{assumption:core-large}
\end{assumption}

The above assumption has two components: the first one relaxes
Assumption~\ref{assumption:core} to accommodate the notion of
$\xi$-incorrectness, and the second one states that, unlike
Assumption~\ref{assumption:core}, there exists a path that not only
does not contain any state-action pairs that are known to be
$\xi$-incorrect, but also that any state-action pair on the path is at
least $\delta$ distance, as measured by the metric $d$, away from any
state-action pair that is known to be $\xi$-incorrect. The second
component makes this assumption stronger. However, it can lead to
substantial speedups in the time taken to reach a goal as we can place
hyperspheres of radius $\delta$ to quickly ``cover'' the
$\xi$-incorrect set.

Algorithm~\ref{alg:large-state-spaces} employs approximate planning by
using a function approximator for cost-to-go estimates and performing
batch updates to fit the approximator. This is necessary as the state
space is large, and maintaining tabular cost-to-go estimates for
each state is expensive in memory and would take a large
number of timesteps to update them in practice. However, for
ease of analysis, we will assume that we do exact updates and maintain
tabular cost-to-go estimates like
Algorithm~\ref{alg:small-state-spaces}. Then, we can show the following
guarantee:
\begin{theorem}
  Assume Assumption~\ref{assumption:core-large} holds then, if
  $\incorrectset^\xi$ denotes the set of all
  $\xi$-incorrect state-action pairs, and the initial cost-to-go estimates
  are admissible and consistent, then using
  Algorithm~\ref{alg:large-state-spaces} with exact updates
  and tabular representation for cost-to-go estimates, the robot is
  guaranteed to
  reach a goal state in at most $|\statespace|^2$
  timesteps. Furthermore, if we allow for $K = |\statespace|$ expansions,
  then we can guarantee that the robot will
  reach a goal state in at most
  $|\statespace|(\covering(\delta) + 1)$ timesteps, where
  $\covering(\delta)$ is the covering number of the set
  $\incorrectset^\xi$.
  \label{thm:large-state-spaces}
\end{theorem}

Proof of the above theorem is given in
Appendix~\ref{sec:proof-theor-refthm:l}. The above theorem states
that, using
Algorithm~\ref{alg:large-state-spaces}, the robot is guaranteed to
reach a goal state, if the initial cost-to-go estimates are admissible
and consistent. The theorem also provides a stronger guarantee that the number of timesteps
to the goal has a dependence on the covering
number, if we do $|\statespace|$ number of expansions at each timestep.
Covering number
$\covering(\delta)$ of a set $A$ is formally defined as the size of
the set $B$ of
state-action pairs $(s, a)$ such that $A \subseteq \bigcup_{(s, a) \in
  B} \mathsf{sphere}(s, a, \delta)$. Note that the covering number
$\covering(\delta)$ is
typically much smaller than the size of the set
$\incorrectset^\xi$. Although performing $|\statespace|$ expansions at
each timestep is infeasible in large state spaces with real-time
constraints, it is useful to note that we achieve speedup from adding
hyperspheres of radius $\delta$. Importantly, the efficiency
of the Algorithm~\ref{alg:large-state-spaces} degrades gracefully with
decreasing $\delta$ and reduces to the bound presented in
Theorem~\ref{thm:small-state-spaces}, if only
Assumption~\ref{assumption:core} holds. Similar to the
worst-case bounds presented in Theorem~\ref{thm:small-state-spaces},
the number of timesteps it takes for the robot to reach a goal state,
in practice as shown in our experiments, has a much smaller dependence on size of state space if
we start with cost-to-go estimates that are reasonably accurate for
the initial model $\hat{M}$, and use cost-to-go
function approximation as we do in
Algorithm~\ref{alg:large-state-spaces}.

\section{Experiments}
\label{sec:experiments}

We test the applicability and efficiency of our approach \textsc{Cmax} on a
range of robotic tasks across simulation and real-world
experiments.\footnote{Code to reproduce simulated
experiments can be found at
\url{https://github.com/vvanirudh/CMAX}}
In simulated experiments, we record the
mean and standard error for the number of timesteps taken by the
robot to reach the goal emphasizing the performance of
\textsc{Cmax}. For physical robot experiments, we present
real-time execution statistics of \textsc{Cmax}.

\subsection{Simulated 4D Planar Pushing in the Presence of Obstacles}
\label{sec:simulated-4d-planar}

\begin{figure}[t]
  \centering
  {\scriptsize
  \begin{subfigure}{0.1\linewidth}
    \begin{tabular}{|c|c|c|}
      \hline
      & \textbf{Steps} & \textbf{\% Success} \\
      \hline
      \textbf{\textsc{Cmax}} & $47 \pm 6$&$ 100\%$ \\
      \hline
      \textbf{RTAA*} & $138 \pm 65 $&$ 30\%$ \\
      \hline
    \end{tabular}
  \end{subfigure}}
  \hspace{40mm}
  \begin{subfigure}{0.3\linewidth}
    \includegraphics[width=\linewidth]{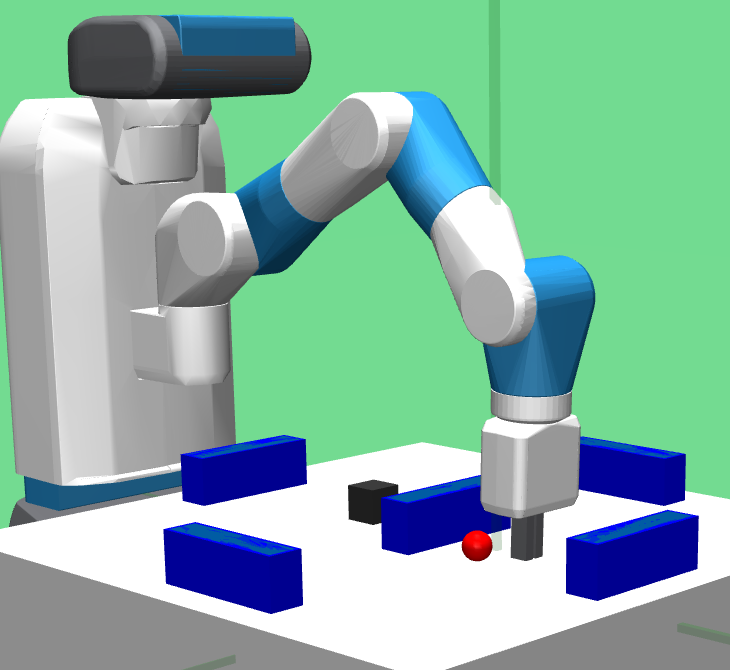}
  \end{subfigure}
  \caption{(left) Results for simulated 7D arm planning experiment
    comparing RTAA* and \textsc{Cmax}. Each entry in the Steps column is obtained using 10
    trials with
    random start configurations and goal locations, and we present
    mean and standard error of number of timesteps it takes the arm to
    reach the goal \textit{among successful trials}. The \% success column
    indicates percentage of successful trials where the arm reached the
    goal in less than $300$ timesteps.(right)
    4D Planar Pushing in the presence of obstacles. The task
    is to push the black box to the red goal using the end-effector.}
  \label{fig:search}
  \vspace{-0.5cm}
\end{figure}

In this experiment, the task is for a robotic gripper to push a cube
from a start location to a goal location in the presence of
static obstacles without any resets, as shown in
Figure~\ref{fig:search} (right). This can be represented as a
planning problem in 4D continuous state space $\statespace$ with any state represented as
the tuple $s = (g_x,
g_y, o_x, o_y)$ where $(g_x, g_y)$ are the xy-coordinates of the
gripper and $(o_x, o_y)$ are the xy-coordinates of the object. The
model $\hat{M}$ used for planning \textit{does not} have the static obstacles and the
robot can only discover the state-action pairs that are affected due
to the obstacles through real world executions. The
action space $\actionspace$ is a discrete set of 4 actions that move
the gripper end-effector in the 4 cardinal directions by a fixed
offset using an IK-based controller. The cost of each transition is
$1$ when the object is not at the goal location, and $0$
otherwise.

\begin{table}[t]
  \centering
  \resizebox{\linewidth}{!}
  {
  {
  \begin{tabular}{|c|c|c|c|c|}
    \hline
    & \multicolumn{2}{c|}{\textbf{Accurate Model}} &
                                                               \multicolumn{2}{c|}{\textbf{Inaccurate
                                                               Model}}
    \\
    \cline{2-5}
    & \textbf{Steps} & \textbf{\% Success} &
                                                    \textbf{Steps}
                          & \textbf{\% Success} \\
    \hline
    \textbf{\textsc{Cmax}} & $63 \pm 22$ & $90\%$ & $192 \pm 40$& $ 80\%$ \\
    \hline
    \textbf{Q-Learning} & $34 \pm 5$ & $90\%$ & $441 \pm 100$ & $45\%$\\
    \hline
    \textbf{Model NN} & $62 \pm 26 $&$90\%$&  $348 \pm 82$&$ 15\%$\\
    \hline
    \textbf{Model KNN} & $106 \pm 34 $&$95\%$& $533 \pm 118 $&$50\%$\\
    \hline
    \hline
    \textbf{Plan with Acc. Model} & $63 \pm 22 $&$90\%$ & $364 \pm 53 $&$85\%$\\
    \hline
  \end{tabular}}}
  \caption{Results for the simulated 4D planar pushing task. First
    column corresponds to the case when the environment has no
    obstacles, and the model is accurate. Second column corresponds to
    when the environment has static obstacles. and model (with no obstacles) is
    inaccurate. Each entry in the Steps subcolumn is obtained using $20$
    random start and goal locations, and we present mean and standard
    error of number of timesteps it takes the robot to reach the
    goal \textit{among successful trials}. The \% success subcolumn
    indicates percentage of successful trials where the robot reached the goal in less
    than 1000 timesteps. The last row corresponds to using the planner
  with an accurate model (the same as the environment.)}
\label{tab:fetch}
\vspace{-0.5cm}
\end{table}

We compare \textsc{Cmax} with
the following baselines: a
model-free Q-learning approach
\cite{DBLP:journals/nature/MnihKSRVBGRFOPB15} that learns from
online executions in environment
and does not use the model $\hat{M}$, and a model learning approach that uses
limited-expansion search for planning but updates a learned residual that
compensates for the discrepancy in dynamics between the model and
environment. The model learning approach is very similar to previous works
that learn residual dynamics models and have been shown to work well in
episodic settings \cite{rastogi2018sample, DBLP:conf/icra/HaY15,
  DBLP:conf/iros/SaverianoYFL17}. We chose two function
approximators for the learned residual dynamics to account for
model learning approaches that use global function
approximators such as neural networks (NN)
\cite{DBLP:conf/nips/JannerFZL19}, and local function approximators
such as K-nearest neighbor regression (KNN)
\cite{DBLP:conf/nips/NouriL08, DBLP:conf/atal/JongS07}. Finally, we compare against
a limited-expansion search planner that uses an accurate model with
the full knowledge about
obstacles to understand the
difficulty of the task. Specific details on the architecture and baseline
parameters can be found in Appendix~\ref{sec:4d-planar-pushing}.

For our implementation, we follow Algorithm~\ref{alg:large-state-spaces}
with euclidean distance metric, $\xi = 0.01$, and $\delta =
0.02$. These values are chosen to capture the discrepancies observed
in the object and gripper position when pushed into an obstacle, and the size of
the obstacles. We use the same values for the model learning KNN
baseline to ensure a fair comparison. The results of our experiments are
presented in Table~\ref{tab:fetch}. We notice that all the approaches
have almost the same performance when both model and environment have
no obstacles (first column). This validates that all the baselines
do well when the model is accurate. However, when the model is
inaccurate (second column), the performance varies across baselines.
Q-learning performs decently well since it relies on the model only
for the initialized Q-values and not during online executions, but as
the task is now more difficult, it solves much fewer trials and is
highly suboptimal. It is
interesting to see that model learning baselines do 
not do as well as one would expect. This can be attributed to the
number of online executions required to learn the correct residual,
which can be prohibitively large. Among 
the two model learning baselines, KNN works better since it requires
fewer samples to learn the residual, while NN requires large amounts of
data. In contrast, \textsc{Cmax} does not seek to learn the true
dynamics and instead is more focused on reaching the goal
quickly. When compared with a planner that uses the accurate model with
obstacles and solves $17$ trials (last row in Table~\ref{tab:fetch}),
our approach solves $16$ trials and achieves the lowest mean 
number of timesteps to reach the goal among all baselines.
We would like to note that the planner with accurate model takes a larger number of timesteps because
we used the same initial cost-to-go estimates as other approaches.
The initial cost-to-go estimates are more accurate for the model
with no obstacles than for the model with obstacles. Hence, it spends a
larger number of timesteps updating cost-to-go estimates.
This experiment shows that by focusing on reaching the goal and not
trying to correct the model 
dynamics, \textsc{Cmax} performs the best and solves the most number of
trials among baselines.

\subsection{3D Pick-and-Place with a Heavy Object}
\label{sec:real-world-3d}

\begin{figure*}[t]
  \centering
  \begin{subfigure}{0.13\linewidth}
    \includegraphics[width=\linewidth]{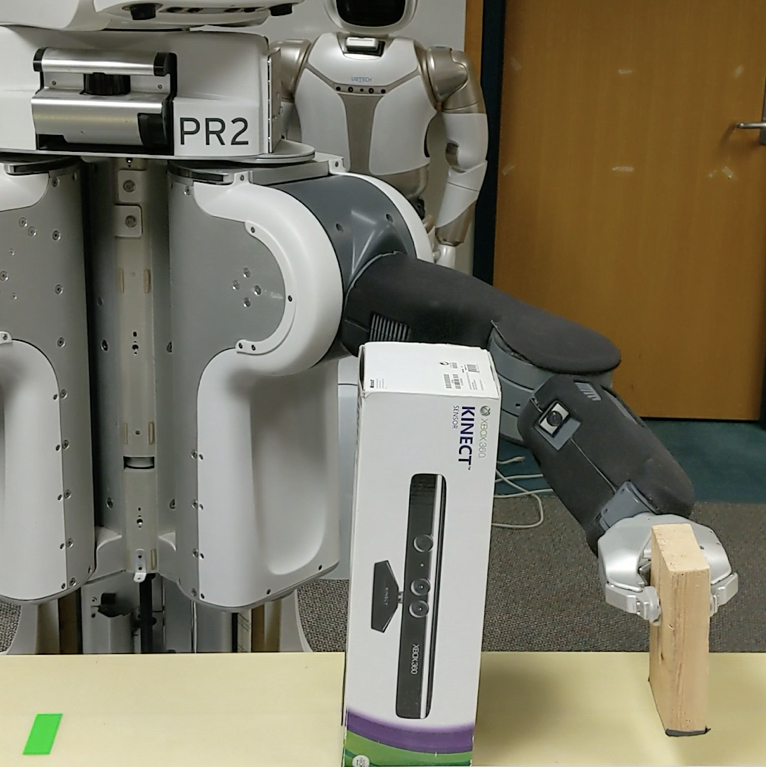}
  \end{subfigure}
  \begin{subfigure}{0.13\linewidth}
    \includegraphics[width=\linewidth]{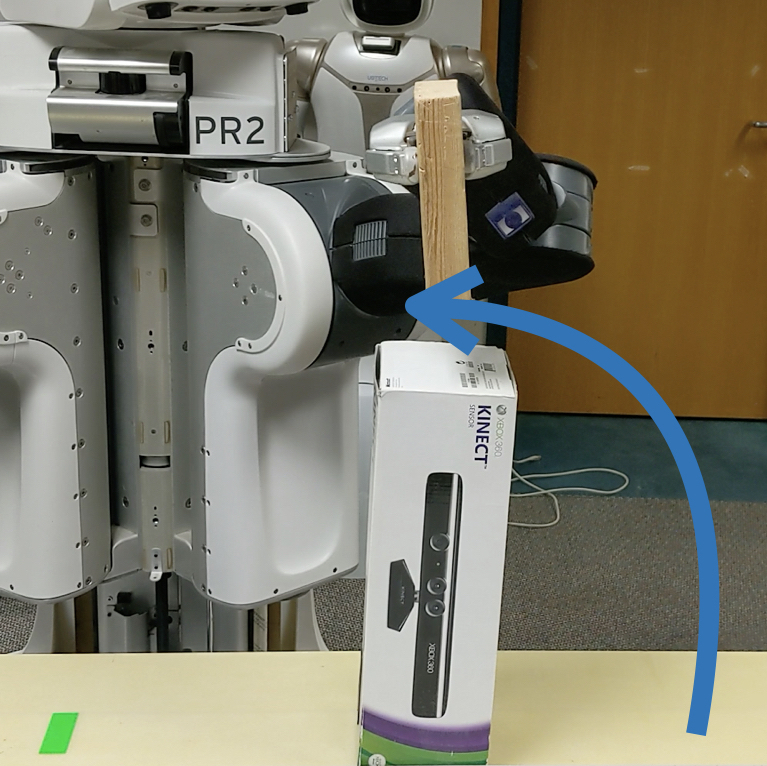}
  \end{subfigure}
  \begin{subfigure}{0.13\linewidth}
    \includegraphics[width=\linewidth]{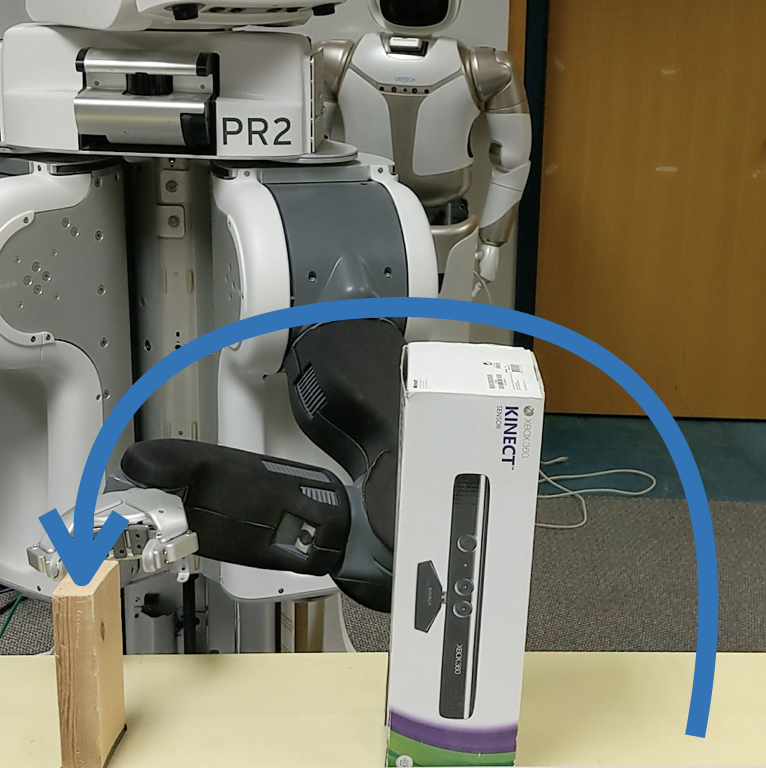}
  \end{subfigure}
  \begin{subfigure}{0.13\linewidth}
    \includegraphics[width=\linewidth]{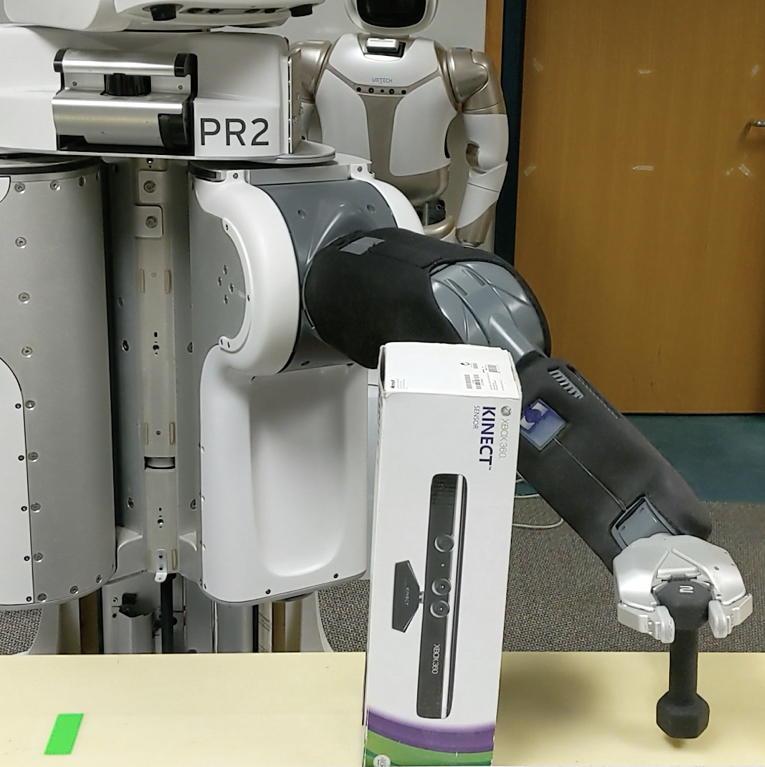}
  \end{subfigure}
  \begin{subfigure}{0.13\linewidth}
    \includegraphics[width=\linewidth]{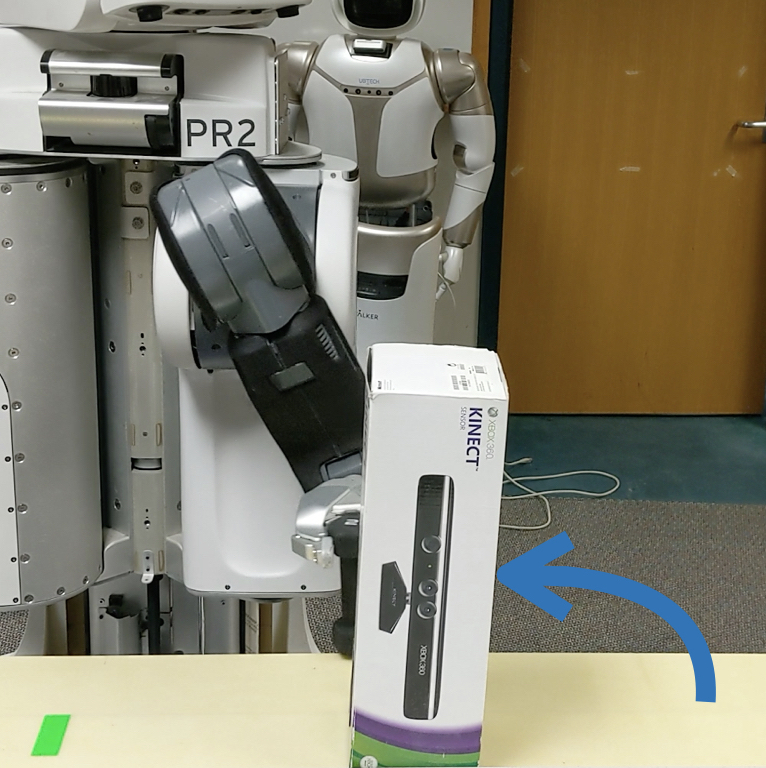}
  \end{subfigure}
  \begin{subfigure}{0.13\linewidth}
    \includegraphics[width=\linewidth]{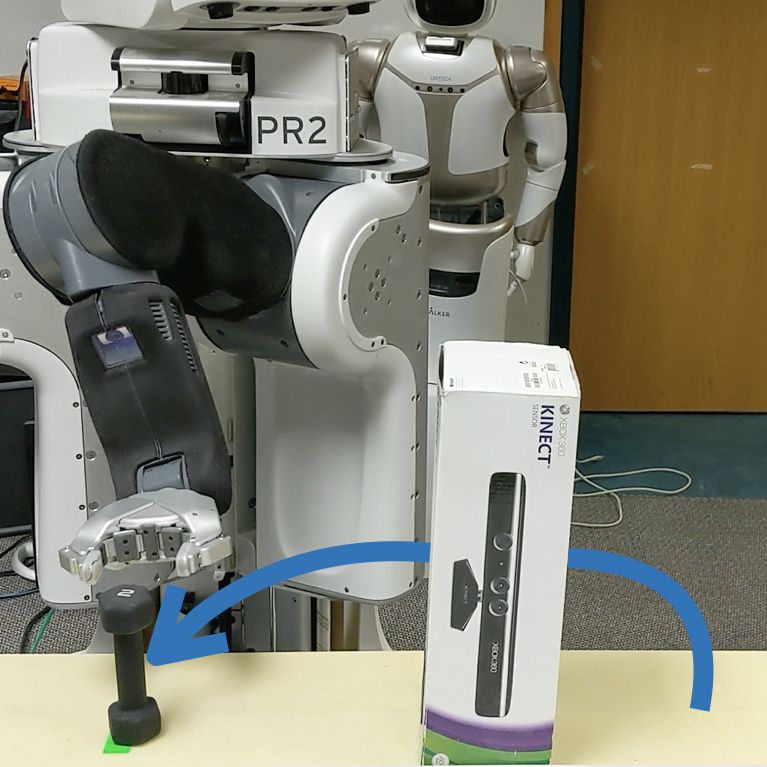}
  \end{subfigure}
  \begin{subfigure}{0.13\linewidth}
    \includegraphics[width=\linewidth]{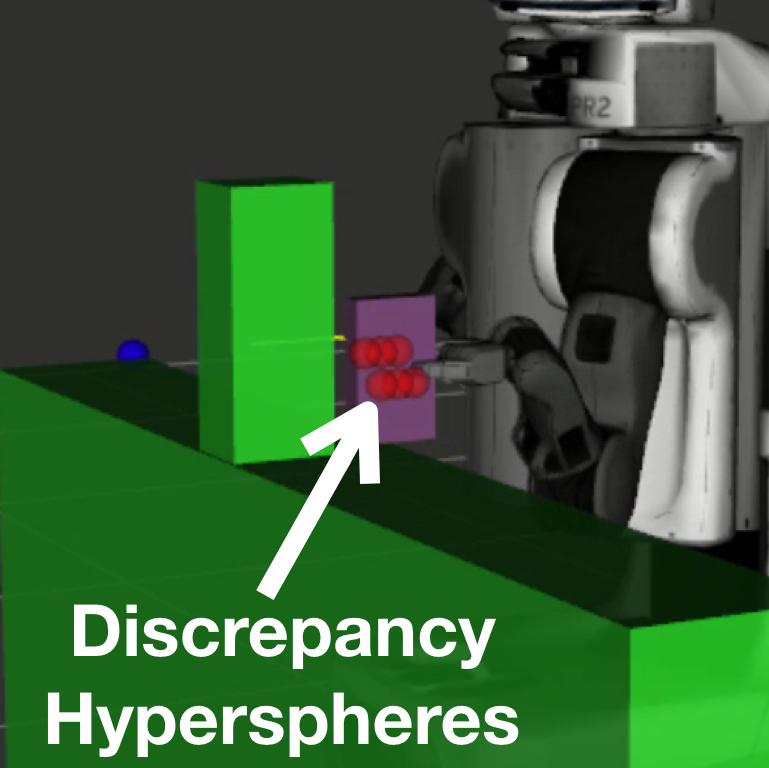}
  \end{subfigure}
  \caption{Physical robot 3D pick-and-place experiment. The task is to
    pick the object (light - wooden block, heavy - black dumbbell) and
    place it at the goal location (green) while avoiding the obstacle
    (box). For the light object (first 3 images), the model dynamics are accurate and
    the robot takes it on the optimal path that goes above the
    obstacle. For the heavy object (next 3 images), the model dynamics are
    inaccurate but using \textsc{Cmax} the robot discovers that there
    is a discrepancy in dynamics when the object is lifted beyond a
    certain height (due to joint torque limits), adds hyperspheres at
    that height to
    account for these transitions (red spheres in the last image), and quickly finds an
    alternate path going behind the obstacle.}
  \label{fig:real-3d}
  \vspace{-0.5cm}
\end{figure*}

The task of this physical robot experiment (Figure~\ref{fig:real-3d}) is to pick and place a heavy 
object using a PR2 arm from a start pick location to a goal place
location while avoiding an obstacle. This can be represented as a
planning problem in 3D discrete state space $\statespace$ where
each state corresponds to the 3D location of the end-effector.
Since it is a
relatively small state space, we use exact planning updates without
any function approximation following
Algorithm~\ref{alg:small-state-spaces} with $K=3$ expansions.
The action space is a
discrete set of $6$ actions corresponding to a fixed offset movement
in positive or negative direction along each dimension.
The
model $\hat{M}$ used by planning \textit{does not} model the object as
heavy and hence, does not capture the dynamics of the arm correctly when it
holds the heavy object. Specific details regarding the experiment can
be found in Appendix~\ref{sec:3d-pick-place}.

We observe that if the object was not heavy, then the arm takes the
object from the start pick location to the goal place location on the
optimal path which goes above the obstacle (first 3 images of
Figure~\ref{fig:real-3d}). However, when executed
with a heavy object, the arm cannot lift the object beyond a certain
height as its joint torque limits are reached. At this point, the robot notes the
discrepancy in dynamics between the model $\hat{M}$ and the real world,
and inflates the cost of any executed transition that tried to move the object
higher. Subsequently, the robot figures out
an alternate path that does not require it to lift the object higher
by taking the object behind the obstacle to the goal place
location (last 4 images of Figure~\ref{fig:real-3d}). The robot takes
36 timesteps (25.8 seconds) to reach
the goal with the heavy object, in comparison to 26 timesteps (22.8 seconds) for the
light object (see video). Thus, the robot using \textsc{Cmax} successfully completes the task despite having a
model with inaccurate dynamics.

\subsection{7D Arm Planning with a Non-Operational Joint}
\label{sec:real-world-7d}

\begin{figure}[t]
  \centering
  \begin{subfigure}{0.3\linewidth}
    \includegraphics[width=\linewidth]{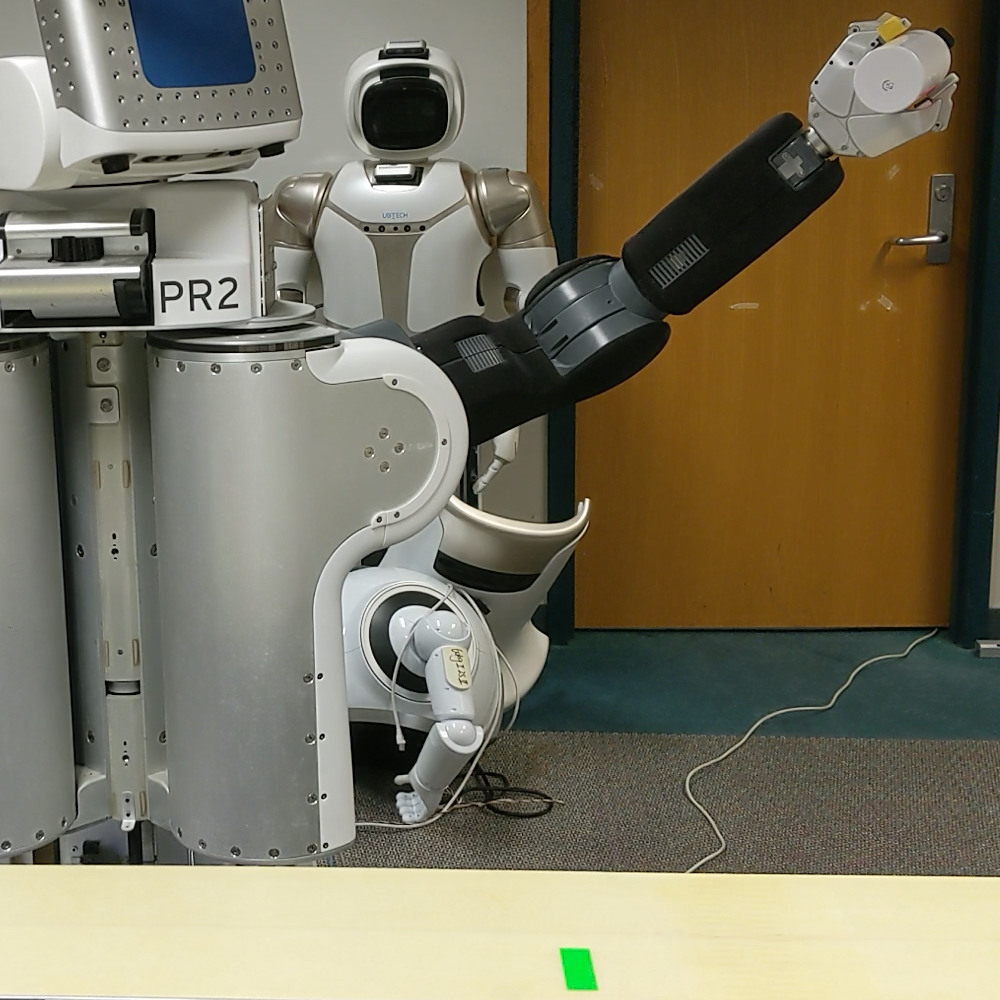}
  \end{subfigure}
  \begin{subfigure}{0.3\linewidth}
    \includegraphics[width=\linewidth]{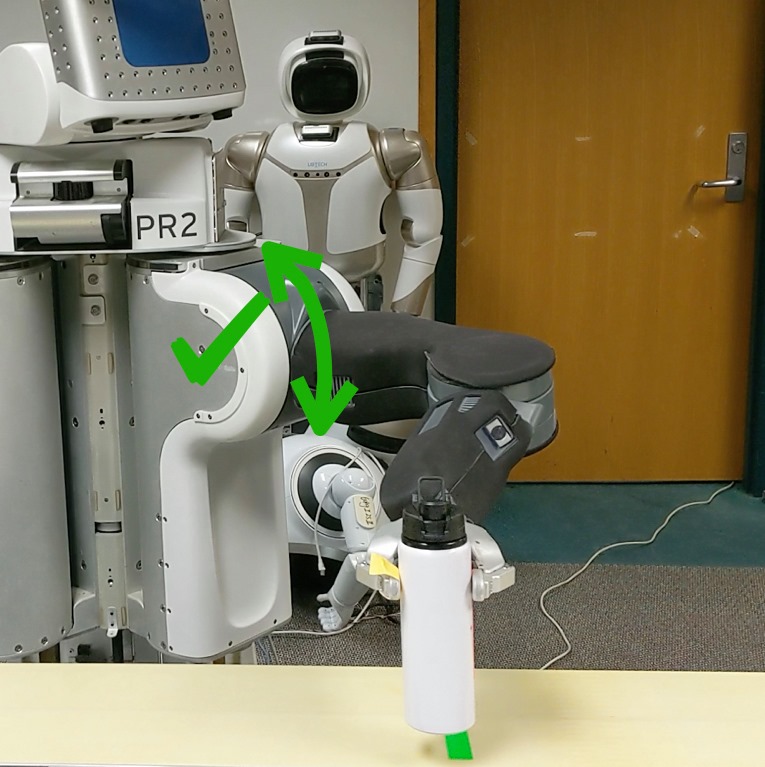}
  \end{subfigure}
  \begin{subfigure}{0.3\linewidth}
    \includegraphics[width=\linewidth]{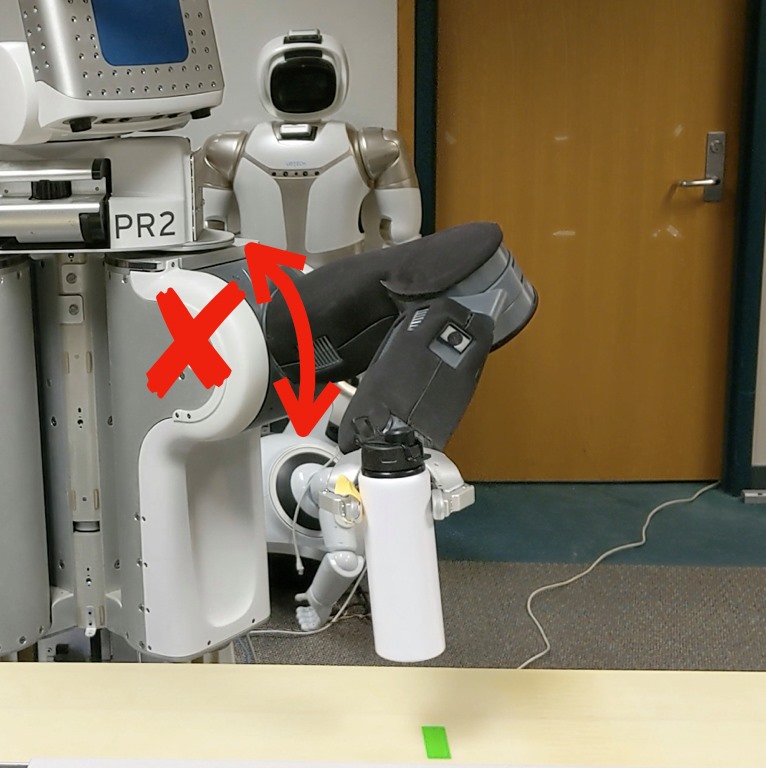}
  \end{subfigure}
  \caption{Physical robot 7D arm planning experiment. The task is to start
  from a fixed configuration (shown in the first image) and move the
  arm so that the end-effector reaches the object place location
  (green). When the shoulder lift joint is operational, the robot uses
  the joint to quickly find a path to the goal (middle image). However, when the
  joint is non-operational, it encounters discrepancies in its model
  and compensates by finding a path that uses other joints to reach
  the goal (last image.)}
\label{fig:real-7d}
\vspace{-0.5cm}
\end{figure}

The task of this physical robot experiment (Figure~\ref{fig:real-7d}) is
to move the PR2 arm with a
non-operational joint from a start configuration so that the
end-effector reaches a goal location, specified as a 3D
region. We represent this as a planning problem in 7D
discrete statespace $\statespace$ where each dimension corresponds to
a joint of the arm bounded by its joint limits. The action space
$\actionspace$ is a discrete set of size 
$14$ corresponding to moving each joint by a fixed offset in the
positive or negative direction. The model $\hat{M}$ used for
planning \textit{does not} know that a joint is non-operational and
assumes that the arm can attain any configuration within the joint
limits. In the real world, if the robot tries to move the
non-operational joint, the arm does not move. Specific details regarding the experiment can
be found in Appendix~\ref{sec:7d-arm-planning}.

For the purpose of this
experiment since the state space is very large, we follow Algorithm~\ref{alg:large-state-spaces} with $\delta
= 1$, $\xi = 1$, and make the shoulder lift joint (marked by red cross and arrows
in last image of Figure~\ref{fig:real-7d}) of PR2 non-operational. We use a
kernel regressor with RBF kernel of length scale $\gamma = 10$ for
the cost-to-go function approximation. Figure~\ref{fig:real-7d} shows \textsc{Cmax} operating in the
real world to place an object at a desired location with a goal
tolerance of $10$ cm. When the shoulder lift joint is operational, the robot finds a
path quickly to the place location by using the joint (middle image of
Figure~\ref{fig:real-7d}). However, when the shoulder lift joint is
non-operational, the robot notes discrepancy in dynamics whenever it tries to
move the joint, places hyperspheres in 7D to inflate the cost, and
comes up with an alternate path (last image of
Figure~\ref{fig:real-7d}) to reach the place location. The 
robot takes $13$ timesteps (32.4 seconds) to reach the goal location with the
non-operational joint, in comparison to $10$ timesteps (25.8 seconds) for the case
where the joint is working (see video). Thus,
the robot successfully finds a path to the place location despite using a
model with inaccurate dynamics.

To emphasize the need for cost-to-go function approximation and
local generalization from hyperspheres in large state spaces, we compared \textsc{Cmax}
against RTAA*, an exact planning method that uses a tabular
representation for cost-to-go estimates and updates model dynamics
online. Results are presented in 
Figure~\ref{fig:search} (left) and show that RTAA* fails to solve $7$
of the $10$ trials whereas \textsc{Cmax} solves all of them, and in
smaller mean number of timesteps.


\subsection{Effect of Function Approximation and Size of Hyperspheres}
\label{sec:underst-effect-funct}
\begin{figure}[t]
  \centering
  \begin{subfigure}{0.45\linewidth}
    \includegraphics[width=\linewidth]{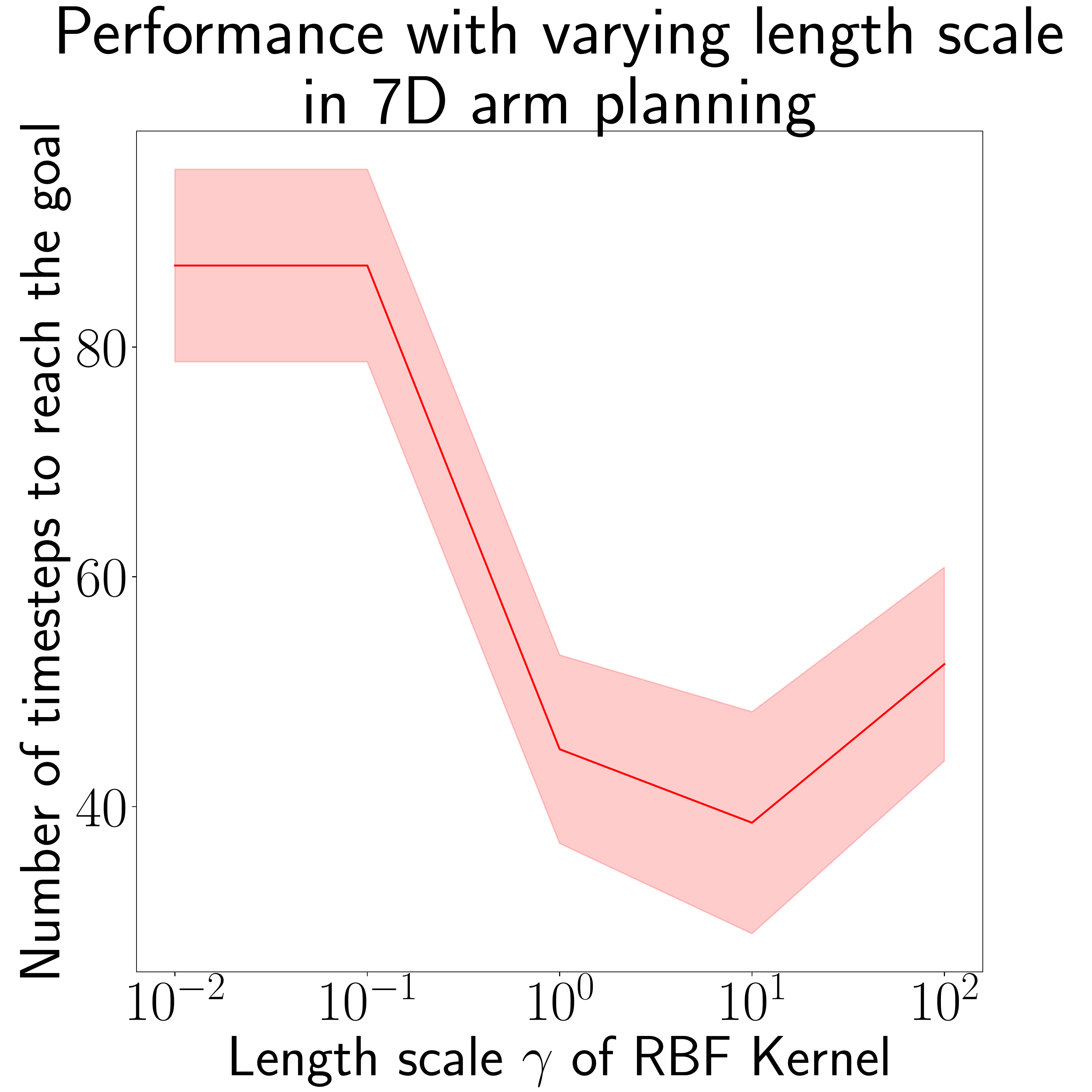}
  \end{subfigure}
  \begin{subfigure}{0.45\linewidth}
    \includegraphics[width=\linewidth]{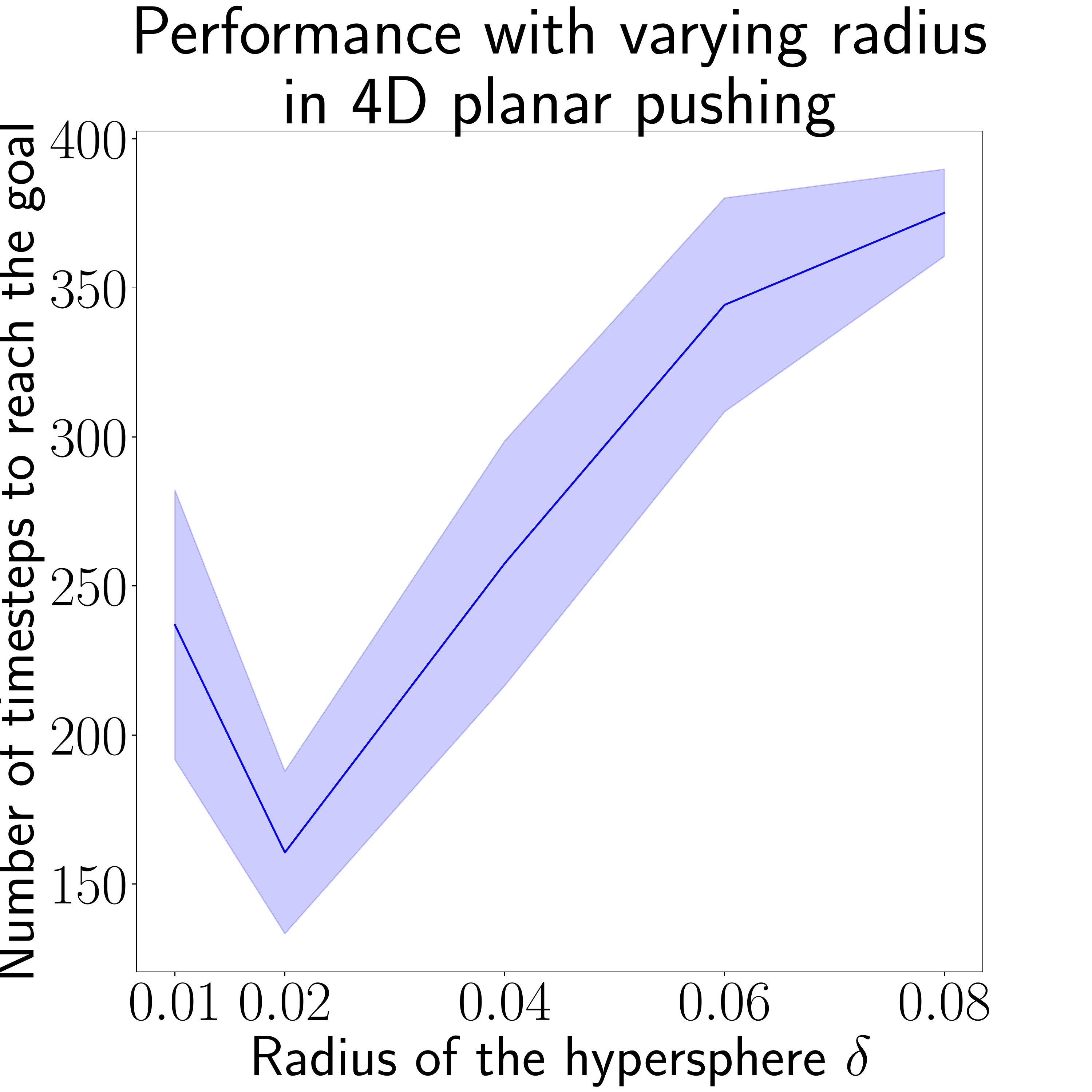}
  \end{subfigure}
  \caption{(left) Performance of \textsc{Cmax} for 7D arm planning as
    the smoothness of the cost-to-go function approximator varies. The plot is
    generated for each value of length scale $\gamma$ by generating
    $10$ random start configurations and goal locations, and running
    our approach for a maximum of $100$ timesteps. (right) Performance
    of our approach for 4D planar pushing as the radius of the
    hypersphere $\delta$ varies. The plot is generated for each value
    of radius $\delta$ by generating $10$ random start and goal
    locations, and running \textsc{Cmax} for a maximum of $400$ timesteps.}
  \label{fig:gamma-radius}
  \vspace{-0.5cm}
\end{figure}

While previous experiments have tested \textsc{Cmax} against other
baselines and on a physical robot, this experiment is designed to
evaluate the effect of cost-to-go function approximation and the size
of hyperspheres on the
performance of \textsc{Cmax} in large state spaces (Algorithm~\ref{alg:large-state-spaces}.)
For the first set of experiments (Figure~\ref{fig:gamma-radius} left), we use the setup of
Section~\ref{sec:real-world-7d} and focus on varying the smoothness of the kernel
regressor cost-to-go function approximation
by varying the length scale $\gamma$ of the RBF kernel.
Intuitively, small length scales result in approximation with high
variance, and for large scales we obtain highly smooth approximation.
We notice that for small $\gamma$, the performance is poor and
as $\gamma$ increases, the performance of \textsc{Cmax}
becomes better as it can generalize the cost-to-go estimates
in the state space. However, for large $\gamma$ the performance
deteriorates as it fails to capture the difference in cost-to-go
values among nearby states due to excessive smoothing. This showcases
the need for generalization in cost-to-go
estimates for efficient updates in large state spaces.

For the second set of experiments (Figure~\ref{fig:gamma-radius}
right), we vary the radius of the 
hyperspheres $\delta$ introduced whenever an incorrect
state-action pair is discovered in
Algorithm~\ref{alg:large-state-spaces}. We use the setup of
Section~\ref{sec:simulated-4d-planar}, vary 
$\delta$ and observe the number of timesteps it takes the robot to
push the object to the goal. We observe that when $\delta$ is
large, the performance is poor as we potentially penalize state-action
pairs that are not incorrect and could result in a very suboptimal
path. However, a very small $\delta$ can also lead to a poor
performance, as we need more online executions to discover the set of
incorrect state-action pairs. Hence, the radius $\delta$ needs to be
chosen carefully to quickly ``cover'' the incorrect set, while not
penalizing any correct state-action pairs.

\subsection{Simulated 2D Gridworld Navigation with Icy States}
\label{sec:simul-2d-gridw}

\begin{table}[t]
  \centering
  \begin{tabular}{|c|c|c|c|}
    \hline
    \textbf{\% Ice}& \textbf{0\%} & \textbf{40\%}
    & \textbf{80\%} \\
    \hline
    \textbf{\textsc{Cmax}} & $78 \pm 4 $ & $231
                                                                \pm
                                                                18
                                                                $
    & $2869 \pm 331 $ \\
    \hline
    \textbf{RTAA*} & $78 \pm 4 $  &
                                                                  $219
                                                                  \pm
                                                                  18
                                                                  $
    & $2185 \pm 249 $ \\
    \hline
    \textbf{Q-Learning} & $3914 \pm 303 $ &
                                                               $1220
                                                               \pm
                                                               103
                                                               $
    & $996 \pm 108 $ \\
    \hline
  \end{tabular}
  \caption{Results for gridworld navigation in presence of
    icy states for a grid of size $100 \times 100$. Each entry is obtained using 50 random seeds, and we
    present the mean and standard error of the number of timesteps it
    takes the robot to reach the goal. The columns represent the
    percentage of icy states in the gridworld.
}
\label{tab:icy}
\vspace{-0.5cm}
\end{table}

In our final experiment, we want to understand the performance of \textsc{Cmax}
compared to other baselines in small domains where model dynamics
can be represented using a table, and can be updated efficiently.
We consider the 2D gridworld such
as the one shown in Figure~\ref{fig:intro}(right) with icy
states where the robot slips (moving left or right on ice moves the
robot by two cells.) The model used for planning
\textit{does not} contain ice, and is an
empty gridworld.
The results are
presented in Table~\ref{tab:icy}. We can observe that model-free approaches like
Q-learning perform well compared to model-based approaches in cases where the model available is
highly inaccurate (see Table~\ref{tab:icy} last column.) However, when
the model is reasonably accurate
RTAA* performs the best. But the
results show that even in domains where model 
dynamics are simple and can be updated efficiently, \textsc{Cmax} competes closely
with RTAA*. Thus, our approach is still applicable in such
domains and is relatively easier to implement.

\section{Related Work}
\label{sec:related-work}

The proposed approach has components concerning real-time heuristic
search, local function approximation methods, and dealing with
inaccuracy in models. There is a wide array of existing work at the
intersection of planning and learning that deal with these
topics. Notably, we leverage prior 
work on real-time heuristic search \cite{DBLP:journals/ai/Korf90,
  DBLP:conf/atal/KoenigL06} for the limited-expansion search-based
planner presented in
Algorithm~\ref{alg:limited-expansion-search}. Using local function
approximation methods in
robotics has been heavily explored in seminal works \cite{
DBLP:conf/icml/VijayakumarS00, DBLP:conf/icra/AtkesonS97} due to their smaller sample complexity
requirements and local generalization properties that do not cause
interference \cite{DBLP:journals/air/AtkesonMS97a,
  DBLP:conf/icml/CoatesAN08}. More recently,
\cite{DBLP:conf/atal/JongS07}, 
\cite{DBLP:conf/nips/NouriL08} and \cite{DBLP:journals/ml/BernsteinS10} have also proposed approaches that
learn local models from online executions. However unlike \textsc{Cmax},
they use these models to approximate the dynamics of the real world. Our work is also closely related to
the field of real-time reinforcement learning that tackles the problem
of acting near-optimally in unknown environments, without any resets~\cite{DBLP:journals/sigart/Sutton91, DBLP:journals/ai/BartoBS95, DBLP:conf/aaai/KoenigS93}. The analysis presented in
Theorem~\ref{thm:small-state-spaces} and \ref{thm:large-state-spaces} borrows several useful results
from \citet{DBLP:conf/aaai/KoenigS93}. Prior
works in model-based reinforcement learning with provable guarantees,
such as \cite{DBLP:journals/ml/KearnsS02, DBLP:journals/ml/BernsteinS10,
  DBLP:journals/jmlr/BrafmanT02, DBLP:conf/icml/KakadeKL03}, are also
related. However, these works learn the true
dynamics by updating the model and give sample complexity results in
the finite-horizon setting or discounted infinite-horizon setting,
unlike our shortest path setting. Among these works,
\citet{DBLP:conf/icml/KakadeKL03}, which proposes a method for exploration in metric
state spaces, serves as an inspiration for the covering number
bounds given in
Theorem~\ref{thm:large-state-spaces}. The work that is most closely
related to ours is \citet{DBLP:conf/aaai/Jiang18}
which proposed an approach that uses a similar idea of updating the cost
function, in cases where updating the model dynamics is infeasible. However, their
approach is suitable only for episodic settings and small state
spaces. Concurrent work by \citet{DBLP:journals/ral/McConachiePMB20}
employs a binary classifier trained on offline data to predict whether
a transition is incorrect or correct, that is then queried during
online motion planning to construct the search tree consisting of only
transitions that are classified as correct.

\section{Discussion and Conclusion}
\label{sec:discussion}
\textsc{Cmax} is the first approach for interleaving planning and
execution that does not require updating 
dynamics, and is guaranteed to reach the goal despite
using an inaccurate dynamical model.
The biggest advantage of \textsc{Cmax} is
that it does not rely on any knowledge of how the model is inaccurate, and
whether it can be updated in real-time.
Hence, it
is broadly applicable in real world robotic tasks with complex inaccurate
models.
In domains where modeling the true dynamics is
intractable, such as deformable manipulation, \textsc{Cmax} can still
be employed to ensure successful execution.
In comparison, approaches that update the model dynamics
online rely on the flexibility of the model to be updated, knowledge
of what is lacking in the model, and a large number of online
executions to correct it. For example, to learn accurate dynamics for a transition
in $N$-D statespace we need at least $N$ samples in the worst case,
whereas our approach needs only $1$ sample to observe a discrepancy
and inflate the cost. The most important shortcomings of \textsc{Cmax}
are
Assumptions~\ref{assumption:core} and \ref{assumption:core-large},
which are hard to verify, and are not satisfied in several real world
robotic
tasks. For
example, consider the task of opening a spring-loaded door 
which is not modeled as loaded. All transitions would have discrepancy in
dynamics, and \textsc{Cmax} as is would fail at completing the task
in a reasonable amount of time. In addition, the hyperparameter
$\delta$ describing the radius of hypersphere needs to be
tuned carefully for each domain which is a limitation of \textsc{Cmax}.

To summarize, we present \textsc{Cmax} for interleaving planning
and execution using inaccurate models that does not require updating
the dynamics of the model, and still provably completes the task. We
propose practical algorithms for both small and large state spaces,
and deploy them successfully in real world robot tasks showing its broad
applicability. In simulation, we 
analyze \textsc{Cmax} and show that it outperforms baselines that
update dynamics online. Future directions include establishing
similar guarantees like Theorem~\ref{thm:large-state-spaces} in the
approximate planning setting, and relaxing
Assumptions~\ref{assumption:core}, \ref{assumption:core-large} so that
\textsc{Cmax} is applicable to a wider range of robotic tasks.

\section*{Acknowledgements}
\label{sec:acknowledgements}

{\small
The authors would like to thank the Search Based Planning Laboratory (SBPL)
for insightful discussions and the anonymous reviewers for their
useful feedback. In addition, the authors would also like to
thank Pragna Mannam, Dhruv Saxena and Allison Del Giorno for their
help in reviewing an initial draft, and Nan Jiang for useful
feedback. AV would like to thank Fahad Islam for 
his help in bringing PR2 to an operational state. AV is supported by
the CMU presidential fellowship endowed by TCS. This work was in part
supported by ONR grant N00014-18-1-2775.
}


\bibliographystyle{plainnat}
\bibliography{allref}
\clearpage
\appendix

\subsection{Proof Sketch of Theorem \ref{thm:small-state-spaces}}
\label{sec:proof-theor-refthm:s}

From \citet{DBLP:conf/atal/KoenigL06} Theorem 3 and
Assumption~\ref{assumption:core}, we have that using
RTAA*, the robot is guaranteed to reach a goal state. Combining this
result with the $|\statespace|^2$ upper bound on the number of timesteps it takes for LRTA*
(which is equivalent to RTAA* with $K=1$ expansion) to reach the goal
from \citet{DBLP:conf/aaai/KoenigS93}, we have that using Algorithm~\ref{alg:small-state-spaces} a
robot is guaranteed to reach a goal state in at most $|\statespace|^2$
timesteps.

To prove the second part, observe that when we do $K = |\statespace|$
expansions at any timestep $t$ in RTAA* and update the cost-to-go, we obtain the optimal
cost-to-go $V^*$ for the penalized model
$\tilde{M}_{\incorrectset_t}$. Once we obtain the optimal cost-to-go,
there will be no further cost-to-go updates in subsequent timesteps
until we either discover an incorrect state-action pair or reach the
goal. Since the number of incorrect $(s, a)$ pairs is
$|\incorrectset|$ and the length of the longest path is bounded
above by $|\statespace|$, using pigeon hole principle we have
that the robot is guaranteed to reach the goal in at most
$|\statespace|(|\incorrectset| + 1)$ timesteps.

\subsection{Proof Sketch of Theorem \ref{thm:large-state-spaces}}
\label{sec:proof-theor-refthm:l}

The proof of the first part of the theorem is very similar to the
proof of Theorem~\ref{thm:small-state-spaces}. It is crucial to notice
that under Assumption~\ref{assumption:core-large}, we will always have
a path from the current state to a goal that has no transition
within a hypersphere. Thus, using RTAA* guarantees we have that using
Algorithm~\ref{alg:large-state-spaces} a robot is guaranteed to reach
a goal state in at most $|\statespace|^2$ timesteps.

To prove the second part, we use a similar pigeon hole principle proof
as Theorem~\ref{thm:small-state-spaces}. However, since we ``cover''
the incorrect set $\incorrectset^\xi$ with hyperspheres, the number of
times we update our heuristic to the optimal cost-to-go of the
corresponding penalized model is equal to the covering number $\covering(\delta)$ of the
$\incorrectset^\xi$, i.e. the number of radius $\delta$ spheres whose
union is a superset of $\incorrectset^\xi$. Thus, with $K =
|\statespace|$ expansions the robot is guaranteed to reach the goal in
at most $|\statespace|(\covering(\delta) + 1)$ timesteps.

\subsection{4D Planar Pushing Experiment Details}
\label{sec:4d-planar-pushing}

In this experiment, the task is for a robotic gripper to push a cube
from a start location to a goal location in the presence of
static obstacles without any resets, as shown in
Figure~\ref{fig:search} (right). This can be represented as a
planning problem in 4D continuous state space $\statespace$ with any state represented as
the tuple $s = (g_x,
g_y, o_x, o_y)$ where $(g_x, g_y)$ are the xy-coordinates of the
gripper and $(o_x, o_y)$ are the xy-coordinates of the object. The
model $\hat{M}$ used for planning \textit{does not} have the static obstacles and the
robot can only discover the state-action pairs that are affected due
to the obstacles through real world executions. The
action space $\actionspace$ is a discrete set of 4 actions that move
the gripper end-effector in the 4 cardinal directions by a fixed
offset using an IK-based controller. The cost of each transition is
$1$ when the object is not at the goal location, and $0$
otherwise.

For all the approaches (except Q-learning), we use
the following neural network architecture for cost-to-go
approximation: a feedforward network with 3 hidden layers each of $64$
units, the network takes as input a $15$D feature representation of the 4D
state $s = (o_x, o_y, g_x, g_y)$ that is constructed as follows:
\begin{itemize}
\item Relative position of the object w.r.t gripper $\frac{\mathbf{o}
    - \mathbf{g}}{\|\mathbf{o} - \mathbf{g}\|_2}$, where $\mathbf{o} =
  (o_x, o_y)$ is the $2$D object position and $\mathbf{g} = (g_x,
  g_y)$ is the $2$D gripper position
\item Distance between position of the object and gripper
  $\|\mathbf{o} - \mathbf{g}\|_2$
\item Relative position of the object w.r.t. goal $\frac{\mathbf{o} -
    \mathbf{t}}{\|\mathbf{o} - \mathbf{t}\|_2}$ where $\mathbf{t} =
  (t_x, t_y)$ is the $2$D goal location
\item Distance between position of the object and goal location
  $\|\mathbf{o} - \mathbf{t}\|_2$
\item Relative position of the gripper w.r.t goal $\frac{\mathbf{g} -
    \mathbf{t}}{\|\mathbf{g} - \mathbf{t}\|_2}$
\item Distance between position of the gripper and goal location
  $\|\mathbf{g} - \mathbf{t}\|_2$
\item Relative position of the object w.r.t center of the table
  $\frac{\mathbf{o} - \mathbf{c}}{\|\mathbf{o} - \mathbf{c}\|_2}$
\item Distance between position of the object and center of the table
  ${\|\mathbf{o} - \mathbf{c}\|_2}$
\item Relative position of the gripper w.r.t center of the table
  $\frac{\mathbf{g} - \mathbf{c}}{\|\mathbf{g} - \mathbf{c}\|_2}$
\item Distance between position of the gripper and center of the table
  ${\|\mathbf{g} - \mathbf{c}\|_2}$
\end{itemize}

The output of the network is a single scalar value representing the
cost-to-go of the input state. We use ReLU activations after each
layer except the last layer. Instead of learning the cost-to-go from
scratch, we start with an initial cost-to-go estimate that is
hardcoded and the neural network function approximator is used to
learn a residual on top of it. The hardcoded initial cost-to-go
estimate is obtained as follows:
\begin{itemize}
\item For the given object position, construct a target position for
  the gripper to go to as follows:
  \begin{itemize}
  \item Get the angle of the vector pointing from the object to the
    goal location: $\theta = \tan^{-1}(\frac{t_x - o_x}{t_y - o_y})$
  \item The target position for gripper is then given by $\mathbf{gt}
    = (o_x - \frac{\sin(\theta)w}{2}, o_y - \frac{\cos(\theta)w}{2})$
    where $w$ is the width of the object
  \end{itemize}
\item We compute the manhattan distance from the gripper to its target
  position $M(\mathbf{g}, \mathbf{gt})$, and from the object to the
  goal location $M(\mathbf{o}, \mathbf{t})$
\item The hardcoded heuristic is obtained as $\hat{V}(s) =
  \frac{M(\mathbf{g}, \mathbf{gt}) + M(\mathbf{o}, \mathbf{t})}{d}$,
  where $d$ is the fixed offset distance the gripper moves for each action
\end{itemize}
The residual cost-to-go function approximator is initialized in such a
way that it outputs $0$ initially for all $s \in \statespace$. We use
a similar residual Q-value function approximator for Q-learning with
the same architecture but that takes as input the above feature
representation and outputs a vector in $\reals^{|\actionspace|}$,
where each element corresponds to the Q-value for that action in the
input state. We also use hardcoded initial Q-values that are
constructed in a similar fashion $\hat{Q}(s, a) = c(s, a) +
\hat{V}(\hat{f}(s, a))$. To ensure a fair comparison across all
baselines, we use the same neural network function approximator for
cost-to-go, and start with the same initial cost-to-go
estimates.

For the model learning baseline that uses Neural network function
approximator, we use a feedforward neural network with $2$ hidden
layers each of $32$ units, the network takes as input the 4D state $s$
and a one-hot encoding of the discrete action $a$ and outputs a 4D
residual vector. The residual vector is added to the next state
predicted by the model $\hat{f}(s, a)$ to get the learned next
state. The loss function used to train the residual is mean
squared loss.

For the model learning baseline that uses KNN function approximator,
we use a radius of $0.02$, and average the next state residual vector observed
for any state within this radius to obtain the prediction for a new
state residual vector. In the same way as above, this residual vector
is added to the next state predicted by the model $\hat{f}(s, a)$ to
obtain the learned next state.

For all the neural network function approximators, we use an Adam
optimizer with learning rate of $0.001$, and an L2 regularization
constant of $0.01$. We use a batch size of $64$ for training all the
neural network function approximators. For Q-learning, we use an
random exploration probability of $\epsilon = 0.1$ and change the
target network by a polyak coefficient of $0.9$.

For all the approaches, we use a
limited expansion search planner with $K = 5$ expansions, $N = 5$
planning updates, batch size $B = 64$, and an Adam optimizer \cite{DBLP:journals/corr/KingmaB14} with
learning rate $\eta = 0.001$.

In training the cost-to-go function approximation, we use the
hindsight experience replay trick with a probability of $0.8$ for
sampling any future state in the trajectory as the desired goal. This
helps in keeping the function approximation stable and also helps in
generalization.

\subsection{3D Pick-and-Place Experiment Details}
\label{sec:3d-pick-place}

The task of this physical robot experiment (Figure~\ref{fig:real-3d})
is to pick and place a heavy object using a PR2 arm from a start pick
location to a goal place 
location while avoiding an obstacle. This can be represented as a
planning problem in 3D discrete state space $\statespace$ where
each state corresponds to the 3D location of the end-effector. In our
experiment, we discretize each dimension into $20$ bins and plan in
the resulting discrete state space of size $20^3$. Since it is a
relatively small state space, we use exact planning updates without
any function approximation following
Algorithm~\ref{alg:small-state-spaces} with $K=3$ expansions. The action space is a
discrete set of $6$ actions corresponding to a fixed offset movement
in positive or negative direction along each dimension. We use a
RRT-based motion planner \cite{DBLP:journals/ijrr/LaValleK01} to plan the path of the
arm between states, while avoiding collision with the obstacle. The
model $\hat{M}$ used by planning \textit{does not} model the object as
heavy and hence, does not capture the dynamics of the arm correctly when it
holds the heavy object. The cost of each transition is $1$ if 
object is not at the goal place location, otherwise it is $0$.

\subsection{7D Arm Planning Experiment Details}
\label{sec:7d-arm-planning}

The task of this physical robot experiment (Figure~\ref{fig:real-7d}) is
to move the PR2 arm with a
non-operational joint from a start configuration so that the
end-effector reaches a goal location, specified as a 3D
region. We represent this as a planning problem in 7D
discrete statespace $\statespace$ where each dimension corresponds to
a joint of the arm bounded by its joint limits. Each dimension is
discretized into $10$ bins resulting in a large state space of
size $10^7$. The action space
$\actionspace$ is a discrete set of size 
$14$ corresponding to moving each joint by a fixed offset in the
positive or negative direction. We use an IK-based controller to
navigate between discrete states. The model $\hat{M}$ used for
planning \textit{does not} know that a joint is non-operational and
assumes that the arm can attain any configuration within the joint
limits. In the real world, if the robot tries to move the
non-operational joint, the arm does not move. Thus, the robot realizes
unreachable states only through real world executions.

\end{document}